\PassOptionsToPackage{unicode}{hyperref}

\PassOptionsToPackage{hyphens}{url}
\documentclass[
]{article}
\usepackage[margin=1in]{geometry}
\usepackage{xcolor}
\usepackage{graphicx}
\usepackage{amsmath,amssymb}
\usepackage{iftex}
\ifPDFTeX
  \usepackage[T1]{fontenc}
  \usepackage[utf8]{inputenc}
  \usepackage{textcomp} 
\else 
  \usepackage{unicode-math} 
  \defaultfontfeatures{Scale=MatchLowercase}
  \defaultfontfeatures[\rmfamily]{Ligatures=TeX,Scale=1}
\fi
\usepackage{lmodern}
\ifPDFTeX\else
\fi
\IfFileExists{upquote.sty}{\usepackage{upquote}}{}
\IfFileExists{microtype.sty}{
  \usepackage[]{microtype}
  \UseMicrotypeSet[protrusion]{basicmath} 
}{}
\makeatletter
\@ifundefined{KOMAClassName}{
  \IfFileExists{parskip.sty}{%
    \usepackage{parskip}
  }{
    \setlength{\parindent}{0pt}
    \setlength{\parskip}{6pt plus 2pt minus 1pt}}
}{
  \KOMAoptions{parskip=half}}
\makeatother
\usepackage[numbers,sort&compress]{natbib}
\usepackage{longtable,booktabs,array}
\usepackage{caption}
\newcolumntype{C}[1]{>{\centering\arraybackslash}p{#1}}
\usepackage{multirow}
\usepackage{calc} 
\usepackage{etoolbox}
\usepackage{tabularx}
\usepackage[colorlinks, backref=page]{hyperref}       

\usepackage{array}
\makeatletter
\patchcmd\longtable{\par}{\if@noskipsec\mbox{}\fi\par}{}{}
\makeatother
\usepackage{bookmark}
\IfFileExists{xurl.sty}{\usepackage{xurl}}{} 
\makeatletter
\@ifundefined{xmpquote}{}{}
\makeatother
\hypersetup{
  pdftitle={Scaling Vision Foundation Models for Toxicologic Pathology},
  hidelinks,
  pdfcreator={LaTeX via pandoc}}

\newenvironment{frontmatterinfo}
  {%
   \par\vspace{0.5em}
   \noindent
   \hspace*{2.5em}%
   \begin{minipage}{\dimexpr\textwidth-5em\relax}
   \small
   \setlength{\parindent}{0pt}
  }
  {%
   \end{minipage}
   \par\vspace{1em}
  }

\title{Scaling Vision Foundation Models for Toxicologic Pathology}
\author{
Jessica~Loo\textsuperscript{1},
Theodoros~Pissas\textsuperscript{2},
Byunghak~Kang\textsuperscript{1},\\
Vincent~A.~Ella\textsuperscript{1},
Xiaoyan~Zhang\textsuperscript{1},
Kevin~Sweeney\textsuperscript{3},
Walter~Georgescu\textsuperscript{2},\\
Prateek~Katiyar\textsuperscript{2},
Julie~Boisclair\textsuperscript{2},
Holger~Hoefling\textsuperscript{2}
\\[1em]
\textsuperscript{1} Novartis Institutes for Biomedical Research Inc., Cambridge, MA, USA\\
\textsuperscript{2} Novartis Pharma AG, Basel, Switzerland\\
\textsuperscript{3} Novartis Ireland Limited, Dublin, Ireland
}
\date{}

\begin{document}
\maketitle

\begin{frontmatterinfo}
\textbf{Funding and Disclosures:} All authors are employees of Novartis.

\vspace{1em}

\textbf{Keywords:} Foundation model, histopathology, non-clinical, preclinical, self-supervised learning, \mbox{toxicology}.

\vspace{1em}

\textbf{Abbreviations:}
\textbf{ABMIL}, attention-based multiple instance learning;
\textbf{CKA}, centered kernel alignment;
\textbf{H\&E}, hematoxylin and eosin;
\textbf{kNN}, k-nearest neighbors;
\textbf{mIOU}, mean intersection-over-union;
\textbf{MMD}, maximum mean discrepancy;
\textbf{MPP}, microns per pixel;
\textbf{PCA}, principal component analysis;
\textbf{PFM}, pathology foundation model;
\textbf{RBF}, radial basis function;
\textbf{TCGA}, The Cancer Genome Atlas;
\textbf{ViT}, Vision Transformer;
\textbf{WSI}, whole slide image.

\vspace{1em}
\end{frontmatterinfo}

\begin{abstract}
\noindent
\normalsize
Toxicology studies to evaluate the safety and toxicity of drug candidates in animal models span the entire drug development pipeline, from early discovery through clinical development and post-marketing, with the preclinical phase representing the most critical milestone before advancement to first-in-human clinical trials.
During each phase, toxicologic pathologists perform systematic histopathology examinations across multiple organs and species, identifying and characterizing morphological changes indicative of drug-induced toxicity.
These workflows are complex and high-volume, requiring substantial manual effort from expert pathologists.
Advances in digital pathology and artificial intelligence are enabling the development of tools and systems that can support and improve pathologists' daily workflows, while pathology foundation models (PFMs) can accelerate and enhance these development efforts by providing robust image representations and flexible computational backbones for a wide range of specialized downstream applications.
However, the current landscape of PFMs remains heavily focused on clinical use cases, leaving critical gaps in the systematic assessment of PFMs in non-clinical settings.
Here, we present TOX-FM, a family of non-clinical PFMs trained on large-scale diverse histopathology datasets of over one million whole slide images consisting of over 60 organs across six animal species from over 14,000 toxicology studies across six decades.
We introduce a comprehensive benchmark of 18 non-clinical downstream tasks, incorporate an established benchmark of 9 clinical downstream tasks, and evaluate TOX-FM against state-of-the-art clinical PFMs across both non-clinical and clinical domains.
Notably, we explore open questions surrounding domain-specific training and cross-domain generalization, offering insights into the opportunities and challenges of applying PFMs in the non-clinical domain.
\end{abstract}

\section{Introduction}\label{introduction}

Toxicology studies are a cornerstone of the drug development pipeline for novel therapeutics.
While the preclinical phase is most critical, providing key evidence to enable the transition from early research to first-in-human clinical trials, toxicology studies are conducted throughout the entire drug development continuum, extending from early discovery through clinical development and post-marketing settings.
At every stage, the safety and toxicity of drug candidates are evaluated in animal models to identify potential adverse effects and to support ongoing risk assessments.
Central to this effort, histopathology examinations across multiple tissues and animal species are performed by toxicologic pathologists in a labor-intensive and time-consuming process, particularly as most tissues examined may not exhibit any drug-related microscopic changes.
The Society of Toxicologic Pathology and other professional groups in pathology and related disciplines have highlighted how artificial intelligence may address some of the current challenges to synergistically improve a pathologist's individual and collective diagnostic efficiency, improving the overall workflow and shortening study timelines \cite{RN1,RN2,RN3,RN4}.

In recent years, many state-of-the-art pathology foundation models (PFMs) have been developed using self-supervised learning frameworks \cite{RN5,RN6,RN7,RN8,RN9,RN10} on large-scale unlabeled datasets of whole slide images (WSIs) \cite{RN11,RN12,RN13,RN14,RN15,RN16,RN17}.
PFMs learn generalizable task-agnostic representations of the underlying data and provide broad support for task-specific machine learning models by enabling efficient adaptation to multiple downstream tasks with relatively limited labeled datasets \cite{RN18}.
The development of PFMs has been driven primarily by clinical applications, particularly oncology, whereas non-clinical PFMs remain comparatively underexplored.
This is partly due to the limited availability of large-scale non-clinical datasets, with most development to date relying on the publicly available TG-GATES dataset, which consists only of rat liver and kidney tissues \cite{RN19,RN20,RN21}.
Additionally, many existing clinical PFMs are subjected to restrictions on commercial use, which may limit their integration into drug development pipelines.
Importantly, the lack of non-clinical PFMs also leaves several important questions unresolved: How critical is domain-specific training for PFMs for downstream task performance?
How well do clinical PFMs generalize to the non-clinical domain, and vice versa?
Addressing these questions is essential for understanding cross-domain generalization and the scope of potential applications of PFMs, as well as whether it is necessary to direct more focus towards the development of non-clinical PFMs.

In this work, we address key gaps in the development and evaluation of foundation models for histopathology by introducing TOX-FM, a family of PFMs trained on large-scale diverse histopathology datasets from animal toxicology studies.
TOX-FM establishes a benchmark for non-clinical PFMs and provides early insights into the cross-domain generalization capabilities of existing clinical PFMs.
Our main contributions are as follows:

\begin{itemize}
\item
  \textbf{A family of large-scale non-clinical PFMs.}
  We present TOX-FM, a family of non-clinical PFMs trained using the DINOv3 self-supervised learning framework \cite{RN22} on over one million WSIs consisting of over 60 organs across six animal species from over 14,000 toxicology studies across six decades.
  To the best of our knowledge, TOX-FM is trained on the largest and most diverse non-clinical histopathology dataset to date and is the first non-clinical PFM to scale beyond the previously limited scope of rat liver and kidney tissues.
\item
  \textbf{A computationally efficient multi-scale dataset curation
  strategy.}
  We use a hierarchical clustering and sampling approach for dataset curation and introduce a progressive multi-scale strategy that operates across magnifications.
  This strategy addresses the practical challenge of scaling training to one million WSIs under a fixed computational budget, while producing a highly diverse yet computationally compact training dataset.
\item
  \textbf{A comprehensive cross-domain benchmark.}
  We introduce a comprehensive benchmark of 18 non-clinical downstream tasks, incorporate an established benchmark of 9 clinical downstream tasks, and evaluate TOX-FM against state-of-the-art clinical PFMs across both non-clinical and clinical domains.
  To the best of our knowledge, this provides the first systematic evaluation of cross-domain generalization of PFMs.
\item
  \textbf{An empirical study of scaling and training strategies.}
  We analyze the impact of scaling model size and dataset diversity, and quantify the contributions of various complementary training
  strategies, including continual learning and Gram anchoring.
  These analyses provide practical and actionable guidance for future development of PFMs.
\end{itemize}

\begin{figure}[!htb]
\centering
\includegraphics[width=\linewidth]{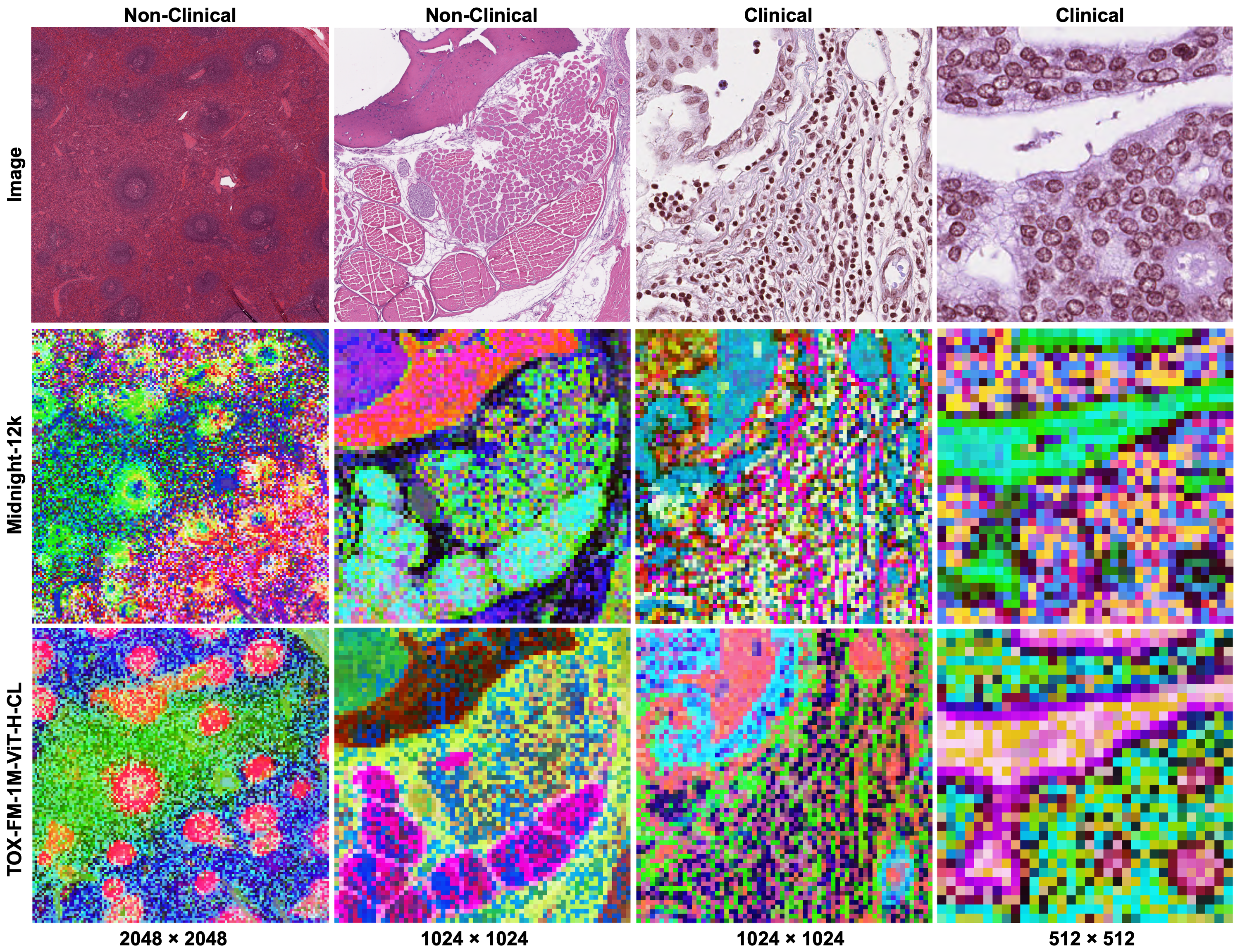}
\caption{Examples of PCA feature maps of Midnight-12k (best performing clinical PFM) and TOX-FM-1M-ViT-H-CL (best performing non-clinical PFM) across different domains and scales.
Non-clinical images are taken from the tissue task (rat, 2.0 mpp) and clinical images are taken from the MoNuSAC task (0.25 mpp).}
\label{fig:figure-1}
\end{figure}

\section{Related Work}\label{related-work}

\textbf{Dataset Scaling and Curation.}
Recent work in the natural vision-language domain has demonstrated that naively increasing the dataset size can be less effective for representation learning, compared to selecting higher quality subsets of available data or more diverse data distributions, especially for similar compute budgets \cite{RN23,RN24,RN25}.
These considerations also apply to large language models, whereby quality-based pruning can retain only a fraction of the available corpus while matching or exceeding the performance of the full dataset \cite{RN26}.
Collectively, these studies suggest that the selection of informative and diverse samples provides a more compute-efficient training signal compared to random sampling, thereby motivating a dataset curation strategy when scaling the dataset to one million WSIs.

\textbf{Generalization Under Domain Shifts.}
To the best of our knowledge, the robustness of PFMs under domain shifts has been studied exclusively in the clinical setting, with prior work focusing on technical factors such as processing and imaging sites \cite{RN27,RN28,RN29,RN30,RN31}, scanner types \cite{RN32,RN33}, and staining conditions \cite{RN34}.
A smaller body of work has examined domain shifts related to biological or disease-relevant factors.
For example, previous work has demonstrated that clinical PFMs trained on hematoxylin and eosin (H\&E) slides from datasets focused on cancer may not generalize well to multi-stain immunohistochemistry slides from datasets focused on autoimmune diseases, with several PFMs performing no better or worse than natural image models \cite{RN35}.
Performance degradation has also been reported for zero-shot cross-cancer evaluations, although the addition of handcrafted biologically motivated descriptors mitigated the loss in performance \cite{RN36}.

Importantly, cross-species domain shifts remain largely unexplored \cite{RN4}.
Some relevant work in this direction includes adversarial training on canine tumors for generalization of a mitosis detection model to human tumors \cite{RN37} and fine-tuning natural image models on rodent liver WSIs \cite{RN38}.
However, the bidirectional generalization of learned image representations between non-clinical animal pathology and clinical human pathology has not been systematically evaluated for PFMs.

\section{Dataset}\label{dataset}

The dataset consisted primarily of H\&E WSIs from toxicology studies conducted by Novartis Biomedical Research across global sites.
The studies consisted of approximately 75\% rodent (rat and mouse) and 25\% non-rodent (dog, monkey, pig, and rabbit) species, ranging from acute (two weeks) to lifelong (two years) carcinogenicity studies.

The training datasets were collected in two phases.
In the first phase, 100,000 WSIs from rat studies conducted between 1999 and 2024 were obtained from existing internal digital databases.
In the second phase, 1 million WSIs were acquired by digitizing archived glass slides from legacy rodent and non-rodent studies conducted from as early as 1967.
We refer to the datasets as \textbf{TOX-FM-100K} and \textbf{TOX-FM-1M}, respectively.
There was no overlap between the studies in the two training datasets.

Overall, \textbf{TOX-FM-100K} was a less diverse dataset, consisting of WSIs from 700 rat studies only, while \textbf{TOX-FM-1M} was a more diverse dataset, consisting of WSIs from 13,352 studies across six species (rat, mouse, dog, monkey, pig, and rabbit).
Additionally, \textbf{TOX-FM-1M} included more studies that were conducted at different contract research organizations, which introduced further variation in staining and scanning protocols, thereby increasing diversity.

\subsection{Dataset Sampling and Curation}\label{dataset-sampling-and-curation}

\begin{figure}[!htb]
\centering
\includegraphics[width=\linewidth]{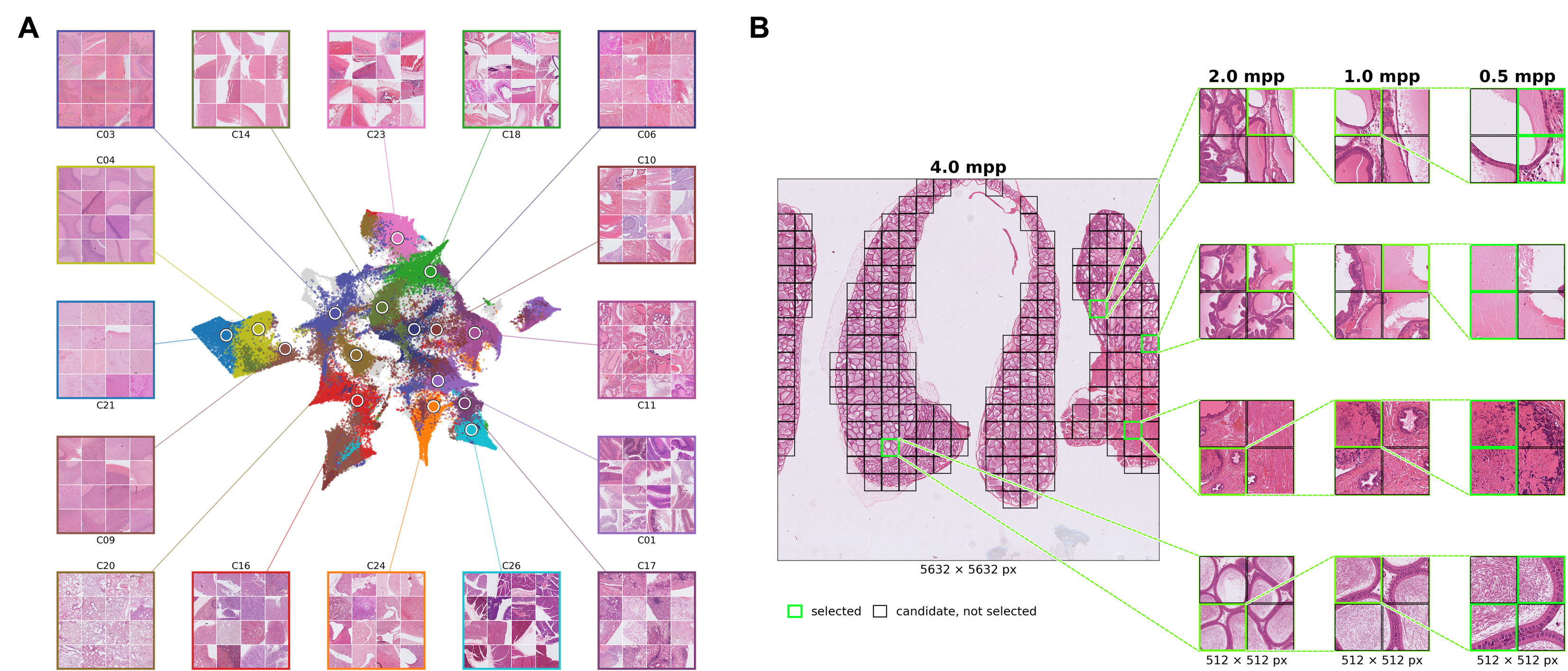}
\caption{\textbf{(A)} UMAP visualization of the hierarchical k-means clustering results at 4.0 mpp.
Each point corresponds to one of the 100,000 bottom-level cluster centroids with the color indicating the corresponding top-level cluster centroid.
Examples of the patches from the 16 largest top-level clusters are shown.
Overall, the hierarchical k-means approach produces a morphologically coherent organization of the embedding space, with top-level clusters forming contiguous regions that correspond to visually distinct histological patterns.
\textbf{(B)} Examples of spatially linked patches selected by the progressive multi-scale strategy from the lowest magnification (4.0 mpp) to highest magnification (0.5 mpp).}
\label{fig:figure-2}
\end{figure}

Patches (256 × 256 pixels) were sampled from the WSIs at multiple magnifications for training.
For \textbf{TOX-FM-100K}, all non-overlapping patches from tissue regions were sampled at 0.5, 1.0, 2.0, and 4.0 microns per pixel (mpp), resulting in approximately 1 billion patches.
For \textbf{TOX-FM-1M}, a dataset curation method was implemented to sample a subset of patches, motivated by recent research on scaling dataset size for vision foundation models \cite{RN10,RN22}, as well as practical considerations such as storage and computational constraints.
The dataset curation resulted in a diverse and compact dataset of approximately 463 million patches, corresponding to approximately 4.4\% of patches selected from a candidate pool of 10.5 billion patches across all magnifications.

We used a progressive multi-scale strategy for sampling patches across the different magnifications, with selection proceeding from the lowest magnification (4.0 mpp) to the highest magnification (0.5 mpp).
At the lowest magnification, all non-overlapping patches from tissue regions from all WSIs were defined as candidate patches for selection.
Candidate patches at each subsequent higher magnification were restricted to those located within the regions of the selected patches at the preceding lower magnification.
\textbf{Figure~\ref{fig:figure-2}B} shows examples of the spatially linked selected patches across magnifications.

We used a hierarchical k-means approach \cite{RN39} for the selection of patches, applying the method to candidate patches at each magnification.
First, Midnight-12k \cite{RN17} was used to generate embeddings for all candidate patches.
The embeddings at each magnification were independently projected into a lower dimensional space using principal component analysis (PCA) to reduce the computational cost of clustering.
Then, k-means clustering was applied to the embeddings and recursively applied to the resulting centroids, yielding a bottom-up four-level hierarchy of clusters.
Given a target subset size, the quota was distributed top-down from the from the root clusters to the leaf clusters, where individual patches were finally selected.
Specifically, at each level, the number of samples allocated to a parent cluster was uniformly distributed among its child clusters.
If a child cluster had fewer samples than the allocated quota, the additional quota was reallocated to the remaining child clusters of the same parent.
\textbf{Figure~\ref{fig:figure-2}A} shows the UMAP visualization of the hierarchical clustering results for the lowest magnification.

Overall, the dataset curation strategy for \textbf{TOX-FM-1M} yielded spatially linked subsets of patches across magnifications.
By balancing the selected patches across the hierarchical levels within a magnification, the method reduced the redundancy of common histological patterns, while preserving diverse and under-represented phenotypes.
Across magnifications, rare tissue types were oversampled relative to the sampling ratio in most cases.
More details of the settings used for the hierarchical k-means approach and a validation analysis of the method are provided in \hyperref[appendix-1]{\textbf{Appendix 1}}.

\section{Training}\label{training}

We used the DINOv3 self-supervised learning framework \cite{RN22} with student-teacher distillation to train Vision Transformer (ViT) models.
We trained ViT-L (305M parameters) and ViT-H (636M parameters) models with the different versions of the datasets to investigate the effects of scaling model size and dataset diversity.
Given the sequential nature of the datasets, which were collected in two phases, we also explored a continual learning strategy to integrate knowledge from both datasets while maintaining efficient use of computational resources.
Additionally, all models were fine-tuned with the Gram anchoring strategy, which was proposed in DINOv3 to improve dense feature representations, to investigate the effects on spatially resolved downstream tasks.

\textbf{Augmentations.}
Models were trained using the multi-crop strategy \cite{RN40}, by applying random cropping scales and using 2 global crops (256 × 256 pixels) and 8 local crops (112 × 112 pixels) per patch.
Geometric augmentations in the form of horizontal and vertical flipping and color augmentations in the form of Gaussian blurring and RandStainNA \cite{RN41} were applied with varying probabilities.
For RandStainNA, the statistical parameters for virtual template generation were estimated from a random subset of 12,000 WSIs from \textbf{TOX-FM-100K}.

\textbf{Optimization.}
Models were trained using 16 or 32 NVIDIA H100 GPUs for ViT-L or ViT-H, respectively.
Patches were sampled with equal probability from each magnification for balanced multi-scale training.
Models were trained for up to 2 million iterations using AdamW optimization \cite{RN42} with a constant learning rate following a linear warmup.

\textbf{Checkpoint Selection.}
12,451 WSIs from 246 studies across all six species were used as a validation dataset to design downstream tasks for model development.
For each model, the average performance across four non-clinical tasks was monitored every 50,000 iterations and the checkpoint with the overall best performance was selected.
Specifically, the validation downstream tasks consisted of species, tissue (rat), artifact, and lesion patch classification using linear probe.
More details of the non-clinical downstream tasks are provided in the \hyperref[evaluation]{\textbf{Evaluation}} section.

\textbf{Continual Learning.}
We explored a continual pre-training strategy \cite{RN43,RN44} whereby the model was initialized with existing weights from training on \textbf{TOX-FM-100K} from the first phase, and training was continued on \textbf{TOX-FM-1M} in the second phase.
The training in the second phase was limited to 1 million iterations.
This reflects a practical real-world scenario whereby foundation models are continuously updated as new datasets become available, with sequential training strategies that preserve previous training efforts and significantly reduce computational costs.
In contrast, joint training strategies, such as retraining from scratch with progressively larger combined datasets, quickly become unsustainable and have been shown to yield diminishing returns relative to continual learning \cite{RN45}.

\textbf{Gram Anchoring.}
The Gram anchoring objective operates on the Gram matrix, which is the matrix of all pairwise dot products of patch features, and aims to push the Gram matrix of the student network towards that of the Gram teacher, which is an earlier checkpoint of the teacher network that typically exhibits superior dense features \cite{RN22}.
For Gram teacher checkpoint selection, we followed the same approach described in the \textbf{Checkpoint Selection} section, but focused only on artifact and lesion tasks, as the quality of the dense features should be best reflected by the performance on these spatially resolved downstream tasks.

\section{Evaluation}\label{evaluation}

\subsection{Pathology Foundation Models}\label{pathology-foundation-models}

We evaluated both non-clinical and clinical PFMs to assess domain-specific performance and cross-domain generalization.
\textbf{Table~\ref{tab:table-1}} shows an overview of the PFMs.

\small
\begin{longtable}{@{}
  >{\centering\arraybackslash}m{(\linewidth - 6\tabcolsep) * \real{0.15}}
  >{\centering\arraybackslash}m{(\linewidth - 6\tabcolsep) * \real{0.35}}
  >{\centering\arraybackslash}m{(\linewidth - 6\tabcolsep) * \real{0.15}}
  >{\centering\arraybackslash}m{(\linewidth - 6\tabcolsep) * \real{0.35}}
@{}}
\toprule
\textbf{Domain} & \textbf{Model} & \textbf{Architecture} & \textbf{Training Magnification (mpp)} \\
\midrule
\endfirsthead
\toprule
\textbf{Domain} & \textbf{Model} & \textbf{Architecture} & \textbf{Training Magnification (mpp)} \\
\midrule
\endhead
\endfoot
\endlastfoot
\multirow{4}{=}{\centering\arraybackslash Non-Clinical}
& TOX-FM-100K-ViT-L & ViT-L & 0.5, 1.0, 2.0, 4.0 \\
& TOX-FM-100K-ViT-H & ViT-H & 0.5, 1.0, 2.0, 4.0 \\
& TOX-FM-1M-ViT-H & ViT-H & 0.5, 1.0, 2.0, 4.0 \\
& TOX-FM-1M-ViT-H-CL & ViT-H & 0.5, 1.0, 2.0, 4.0 \\
\midrule
\multirow{3}{=}{\centering\arraybackslash Clinical}
& Prov-GigaPath & ViT-g & 0.5 \\
& H-optimus-0 & ViT-g & 0.5 \\
& Midnight-12k & ViT-g & 0.25, 0.5, 1.0, 2.0 \\
\bottomrule
\caption{Overview of non-clinical and clinical PFMs for evaluation.
The number of parameters for ViT-L, ViT-H, and ViT-g architectures are approximately 305M, 636M, and 1.1B parameters, respectively.}
\label{tab:table-1} \\
\end{longtable}
\normalsize

\textbf{Non-Clinical Models.}
First, we developed TOX-FM-100K-ViT-L and TOX-FM-100K-ViT-H to evaluate the effects of scaling model size.
While most state-of-the-art PFMs adopt ViT-g (1.1B parameters) architectures, we explored whether ViT-L and ViT-H architectures, which are approximately 30 -- 60\% of the size of ViT-g, remain competitive with significantly reduced computational costs.
Second, we developed TOX-FM-1M-ViT-H to evaluate the effects of scaling dataset diversity and training with curated datasets, as well as to provide a baseline for the continual learning strategy.
Finally, we developed TOX-FM-1M-ViT-H-CL, which was initialized from TOX-FM-100K-ViT-H, to evaluate the continual learning strategy used to integrate knowledge from datasets collected in different phases.
Additionally, we fine-tuned all models with the Gram anchoring objective on their respective training datasets and refer to the corresponding models with the -Gram suffix.

\textbf{Clinical Models.}
We evaluated state-of-the-art clinical PFMs with licenses that permit commercial use.
Specifically, we evaluated Prov-GigaPath \cite{RN12}, H-optimus-0 \cite{RN13}, and Midnight-12k \cite{RN17}, which were trained on various public and proprietary clinical datasets and have demonstrated strong performance across various clinical pathology benchmarks \cite{RN46,RN47,RN48}.
All models were accessed via the official versions published on Hugging Face.

\textbf{Natural Image Models.}
As baselines, we included out-of-domain models trained on non-pathology datasets.
Specifically, we evaluated a ViT-B (86M parameters) model trained on the ImageNet dataset of 1.2 million images \cite{RN49, RN50} and the ViT-L variant (distilled from ViT-7B) of the original DINOv3 model trained on the LVD-1689M web-scale dataset of 1.689 billion images \cite{RN22}.
Although web-scale data may contain some biomedical or pathology-related images, such content is expected to represent only a small fraction of LVD-1689M, which is predominantly composed of natural images.
Accordingly, ImageNet-ViT-B provides a lower bound baseline of a small natural image model, whereas DINOv3-LVD-1689M-ViT-L provides a strong baseline of an architecture-matched general vision foundation model.

\subsection{Downstream Tasks}\label{downstream-tasks}

We evaluated the PFMs outlined in \textbf{Table~\ref{tab:table-1}} on a comprehensive set of 18 non-clinical and 9 clinical downstream tasks across various species, indications, tissues, and magnifications.
The types of tasks included patch classification, slide classification, and segmentation.
\textbf{Table~\ref{tab:table-2}} shows an overview of the downstream tasks.

\small
\begin{longtable}{@{}
  >{\centering\arraybackslash}m{(\linewidth - 8\tabcolsep) * \real{0.15}}
  >{\centering\arraybackslash}m{(\linewidth - 8\tabcolsep) * \real{0.25}}
  >{\centering\arraybackslash}m{(\linewidth - 8\tabcolsep) * \real{0.25}}
  >{\centering\arraybackslash}m{(\linewidth - 8\tabcolsep) * \real{0.18}}
  >{\centering\arraybackslash}m{(\linewidth - 8\tabcolsep) * \real{0.17}}
@{}}
\toprule
\textbf{Domain} &
\textbf{Type} &
\textbf{Task} &
\textbf{Magnification (mpp)} &
\textbf{Number of Classes} \\
\midrule
\endhead
\endlastfoot
\multirow{18}{=}{\centering\arraybackslash Non-Clinical} &
\multirow{8}{=}{\centering\arraybackslash Patch Classification (Linear Probe)}
& Species & 4.0 & 6 \\
& & Tissue (Dog) & 2.0 & 23 \\
& & Tissue (Monkey) & 2.0 & 10 \\
& & Tissue (Mouse) & 2.0 & 33 \\
& & Tissue (Rabbit) & 2.0 & 9 \\
& & Tissue (Rat) & 2.0 & 30 \\
& & Artifact & 1.0 & 9 \\
& & Lesion & 0.5 & 32 \\
\cmidrule(lr){2-5}
& \multirow{2}{=}{\centering\arraybackslash Patch Classification (Attention Probe)}
& Artifact & 1.0 & 9 \\
& & Lesion & 0.5 & 32 \\
\cmidrule(lr){2-5}
& \multirow{6}{=}{\centering\arraybackslash Slide Classification}
& Species & 4.0 & 6 \\
& & Tissue (Dog) & 2.0 & 23 \\
& & Tissue (Monkey) & 2.0 & 10 \\
& & Tissue (Mouse) & 2.0 & 33 \\
& & Tissue (Rabbit) & 2.0 & 9 \\
& & Tissue (Rat) & 2.0 & 30 \\
\cmidrule(lr){2-5}
& \multirow{2}{=}{\centering\arraybackslash Segmentation}
& Artifact & 1.0 & 10 \\
& & Lesion & 0.5 & 33 \\
\midrule
\multirow{9}{=}{\centering\arraybackslash Clinical} &
\multirow{5}{=}{\centering\arraybackslash Patch Classification}
& BACH & 0.5 & 4 \\
& & BreakHis & 0.25 & 4 \\
& & CRC & 0.5 & 9 \\
& & GleasonArvaniti & 0.25 & 4 \\
& & PatchCamelyon & 1.0 & 2 \\
\cmidrule(lr){2-5}
& \multirow{2}{=}{\centering\arraybackslash Slide Classification}
& Camelyon16 (Small) & 0.25 & 2 \\
& & PANDA (Small) & 0.5 & 6 \\
\cmidrule(lr){2-5}
& \multirow{2}{=}{\centering\arraybackslash Segmentation}
& BCSS & 0.25 & 22 \\
& & MoNuSAC & 0.25 & 5 \\
\bottomrule
\caption{Overview of non-clinical and clinical downstream tasks for evaluation of PFMs.}
\label{tab:table-2}
\end{longtable}
\normalsize

\textbf{Non-Clinical Downstream Tasks.}
19,935 WSIs from 289 studies across all six species were used as an evaluation dataset to design downstream tasks for model evaluation.
The evaluation dataset was completely independent and there was no overlap with the WSIs in the \textbf{TOX-FM-100K} and \textbf{TOX-FM-1M} training datasets.
The WSIs were paired with either metadata containing species and tissue information or pathologists' annotations for artifacts and lesions, which were used to obtain labels for the respective tasks.
The pathologists' annotations were constrained to a region of interest in the WSI and exhaustively annotated within the region using QuPath open-source software for digital pathology image analysis \cite{RN51}.
Annotated artifacts were related to slide quality issues caused during slide preparation or scanning, while annotated lesions were related to drug- or disease-induced pathological findings in the tissue.

The non-clinical tasks were designed to encompass a range of task-appropriate magnifications.
Species and tissue tasks were performed at lower magnifications, specifically at 4.0 and 2.0 mpp, respectively, as they are primarily associated with global semantic features of the image.
Artifact and lesion tasks were performed at higher magnifications, specifically at 1.0 and 0.5 mpp, respectively, as they are primarily associated with local spatial features of the image.
\textbf{Figures~\ref{fig:figure-3}B} and \textbf{\ref{fig:figure-3}C} show some examples of the artifacts and lesions present in WSIs.

Following the established best practices for assessing PFMs \cite{RN11,RN13,RN14,RN15,RN16,RN46}, we designed all tasks as patch classification tasks using linear probe evaluation.
For improved performance on artifact and lesion tasks, we also applied attention probe evaluation using attention-based multiple instance learning (ABMIL) \cite{RN52} to aggregate patch token embeddings for patch classification.
For improved performance on species and tissue tasks, we also designed slide classification tasks using ABMIL to aggregate embeddings across multiple patches sampled from a WSI.
Finally, we designed artifact and lesion segmentation tasks using a simple U-Net decoder \cite{RN53}.

For each task, the relevant WSIs were split into training and testing datasets, and patches (256 × 256 pixels) were sampled accordingly from the WSIs.
Classification tasks were designed to be balanced across classes.
To enable fair comparison between models, we performed a hyperparameter sweep across different learning rates and weight decays where applicable.
We report the performance from the best hyperparameter combination, which was determined by monitoring performance on a constant batch that was held out from the training dataset for the task.
We report the macro F1 metric for classification tasks and the mean intersection-over-union (mIOU) metric for segmentation tasks.

All PFMs were frozen for evaluation and only the respective probes were trained for the downstream tasks.
All linear probe patch classification tasks and slide classification tasks used the concatenation of the CLS token and mean of the patch token embeddings of the ViT model.
All attention probe patch classification tasks and segmentation tasks used the patch token embeddings of the ViT model.

\textbf{Clinical Downstream Tasks.}
We adopted the \emph{eva} open-source foundation model evaluation framework for oncology \cite{RN46} and included all accessible pathology datasets.
We followed all evaluation protocols and metrics as defined in the official \emph{eva} repository \cite{RN54}.

Note that the \emph{eva} evaluation protocol specifies constant default hyperparameters.
To enable fair comparison between models, we performed a hyperparameter sweep across different learning rates and weight decays where applicable.
We report the performance from the best hyperparameter combination, similar to the evaluation protocol described for non-clinical tasks.
Overall, this improved performance by approximately 1 -- 4\% compared to the metrics reported on the official leaderboard for the clinical PFMs.

More details of the hyperparameters are provided in \hyperref[appendix-2]{\textbf{Appendix 2}}.

\section{Results}\label{results}

\subsection{Performance of Non-Clinical and Clinical PFMs}\label{performance-of-non-clinical-and-clinical-pfms}

\begin{figure}[!htb]
\centering
\includegraphics[width=\linewidth]{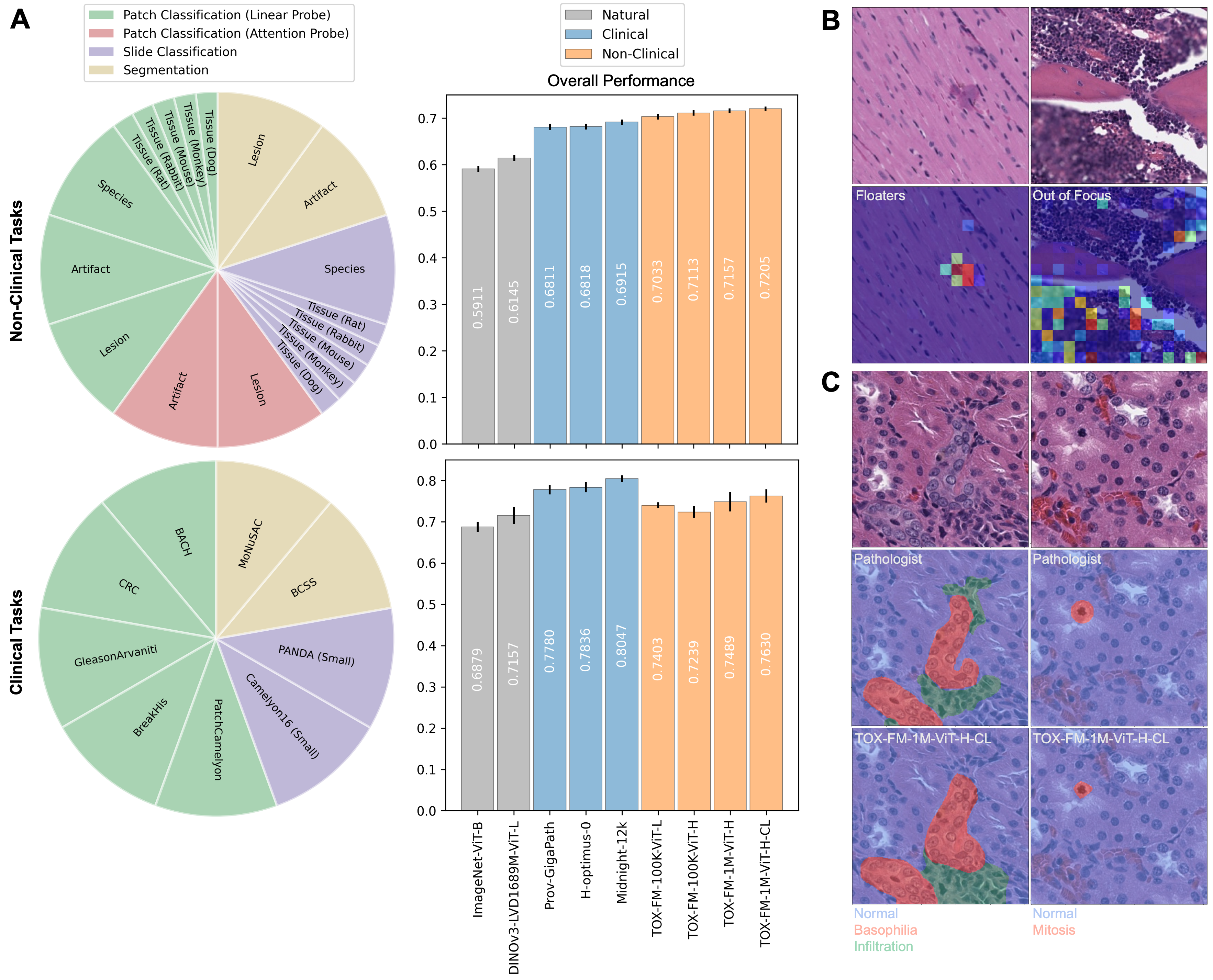}
\caption{\textbf{(A)} Distribution of task weights and overall performance (mean and standard deviation) of all models across a comprehensive set of 18 non-clinical and 9 clinical downstream tasks.
\textbf{(B)} Examples of attention weights from attention probe evaluation for the artifact patch classification non-clinical task using embeddings from TOX-FM-1M-ViT-H-CL.
The attention weights were used to aggregate the patch token embeddings for patch classification.
\textbf{(C)} Examples of segmentation masks for the lesion segmentation non-clinical task using embeddings from TOX-FM-1M-ViT-H-CL.}
\label{fig:figure-3}
\end{figure}

\textbf{Figure~\ref{fig:figure-3}A} shows the overall performance of all models across the non-clinical and clinical downstream tasks.
More details of the performance on individual tasks are provided in \hyperref[appendix-3]{\textbf{Appendix 3}}.

Overall, domain-specific models achieve superior performance on their respective domain-specific tasks, with TOX-FM-1M-ViT-H-CL demonstrating the best performance across the non-clinical PFMs and Midnight-12k demonstrating the best performance across the clinical PFMs.
Notably, cross-domain performance remains remarkably competitive across both non-clinical and clinical PFMs.
Specifically, the best cross-domain model achieves approximately 95\% relative performance compared to the strongest domain-specific model and 98\% relative performance compared to the weakest domain-specific model, suggesting effective cross-domain generalization.

\textbf{Figure~\ref{fig:figure-1}} shows examples of the feature maps of Midnight-12k and TOX-FM-1M-ViT-H-CL across different domains and scales using PCA visualization techniques, whereby the features are projected into 3 dimensions and visualized in RGB space \cite{RN22}.
Across both non-clinical and clinical images, TOX-FM-1M-ViT-H-CL produces high-quality feature maps, capturing the semantic similarity of tissue morphology at different spatial locations across the images.
Similarly, Midnight-12k also generates structured and informative feature maps, although some examples appear less spatially coherent by visual inspection.
However, PCA feature map visualizations are qualitative and projection-dependent, so visual coherence is not necessarily a direct measure of downstream utility or predictive accuracy.
We note that Midnight-12k was trained using the DINOv2 framework which uses learnable positional embeddings, which are interpolated to process images at varying resolutions.
In contrast, TOX-FM models were trained using rotary positional embeddings \cite{RN55}, which enables the models to process images at varying resolutions without requiring adaptation.
Therefore, while the PCA feature maps in \textbf{Figure~\ref{fig:figure-1}} are visualized at different resolutions to support qualitative analysis across multiple scales, additional examples of the PCA feature maps at the training resolution of 256 × 256 pixels are provided in \hyperref[appendix-3]{\textbf{Appendix 3}}.

Across the non-clinical tasks, we observe that the clinical PFMs tend to show the biggest gaps in performance on species and tissue tasks, which are performed at lower magnifications, while the performance remains more competitive on artifact and lesion tasks, which are performed at higher magnifications.
Lower magnifications capture tissue-level architecture and organization, which exhibit greater differences between human and animal tissues, and may have lower transferability from the clinical to non-clinical domain.
At higher magnifications, the distinctions between human and animal tissues are progressively reduced as cell-level details become the focus, resulting in stronger cross-domain transferability.
We note that Prov-GigaPath and H-optimus-0 were trained exclusively on a single magnification at 0.5 mpp.
While the underlying self-supervised learning frameworks \cite{RN7,RN10} do implicitly learn multi-scale features through global and local cropping and scaling augmentations, the effective magnification range remains limited, and such augmentations alone are insufficient to capture the intrinsic multi-scale nature of WSIs.
In contrast, models that explicitly include multi-scale datasets sampled at various magnifications such as Midnight-12k and TOX-FM models demonstrate measurably improved performance, particularly when evaluated across broader magnification ranges.

While linear probe evaluation remains a strong baseline for assessing PFMs, attention probe evaluation significantly improves performance for spatially resolved patch classification tasks.
Specifically, using attention probe evaluation, Midnight-12k demonstrates a relative performance improvement of 18\% and 10\% on artifact and lesion tasks, respectively.
TOX-FM-1M-ViT-H-CL demonstrates an even larger relative performance improvement of 27\% and 13\%, respectively, suggesting that the patch token embeddings encode strong representations which can be appropriately leveraged using attention probe evaluation, but may be diminished when simply averaged for linear probe evaluation.
\textbf{Figure~\ref{fig:figure-3}B} shows examples of the attention weights from attention probe evaluation using embeddings from TOX-FM-1M-ViT-H-CL, whereby the relative importance of the patch token embeddings correlates well with the location of the artifacts.

\subsection{Model Size}\label{model-size}

Our experiments on scaling model size provide further insights into the performance characteristics of PFMs.
First, scaling model size without dataset diversity appears to lead to overfitting and degraded cross-domain performance.
Specifically, when scaling from TOX-FM-100K-ViT-L to TOX-FM-100K-ViT-H, we observe that overall performance improves on non-clinical tasks but decreases on clinical tasks.
However, jointly scaling model size and dataset diversity, as in TOX-FM-1M-ViT-H, improves the overall performance across both sets of tasks.
These observations also highlight the importance of broad and comprehensive evaluations for accurately characterizing the scope and limitations of PFMs.
Without assessing the performance on clinical tasks, it may have been incorrectly assumed that scaling model size universally improves downstream performance, while the effects of overfitting would have remained unrecognized.

We also found that the smaller model size of the non-clinical PFMs compared to the clinical PFMs does not appear to constrain performance on downstream tasks at this stage.
Specifically, despite TOX-FM-1M-ViT-H-CL being 60\% of the size of Midnight-12k, both models exhibit comparable cross-domain performance as detailed above.
Alternatively, if model capacity was a bottleneck, we would expect non-clinical PFMs to perform substantially worse on clinical tasks than clinical PFMs do on non-clinical tasks, yet this asymmetry is not observed.

\subsection{Dataset Diversity}\label{dataset-diversity}

Our experiments on scaling dataset diversity demonstrate a difference in the impact on domain-specific and cross-domain downstream tasks.
Specifically, scaling from TOX-FM-100K-ViT-H to TOX-FM-1M-ViT-H yielded more pronounced performance gains on clinical tasks than on non-clinical tasks.
Motivated by observations from the natural image domain, we hypothesize that this is related to the difference in distributional proximity between the training and task datasets \cite{RN56}.
Under this assumption, \textbf{TOX-FM-100K}, despite being limited to rat studies, may already provide sufficient task-relevant diversity and effective representations for many of the non-clinical tasks, for which rat was the dominant species.
Moreover, the pathologists' annotations for artifacts and lesions were primarily performed on rat WSIs from recent studies, so the additional diversity across species, studies, and variation in staining and scanning protocols offered by \textbf{TOX-FM-1M} may be less relevant for these tasks.

We performed a quantitative analysis to validate the assumption that \textbf{TOX-FM-100K} is distributionally closer to the non-clinical tasks than \textbf{TOX-FM-1M}.
First, to make the analysis computationally tractable, random subsets of 100,000 patches were generated from each of \textbf{TOX-FM-100K}, \textbf{TOX-FM-1M}, the non-clinical task datasets, and the clinical task datasets.
To ensure that each subset was representative of the complete dataset, subsets were stratified across magnifications, tasks, and classes where applicable.
Next, following the standard practice for measuring distances between data distributions \cite{RN57}, we embedded each subset and calculated the maximum mean discrepancy (MMD) with a radial basis function (RBF) kernel \cite{RN58}, sliced Wasserstein distance \cite{RN59}, and Fréchet distance \cite{RN60} between the training and task subsets in the embedding space.

\begin{figure}[!htb]
\centering
\includegraphics[width=\linewidth]{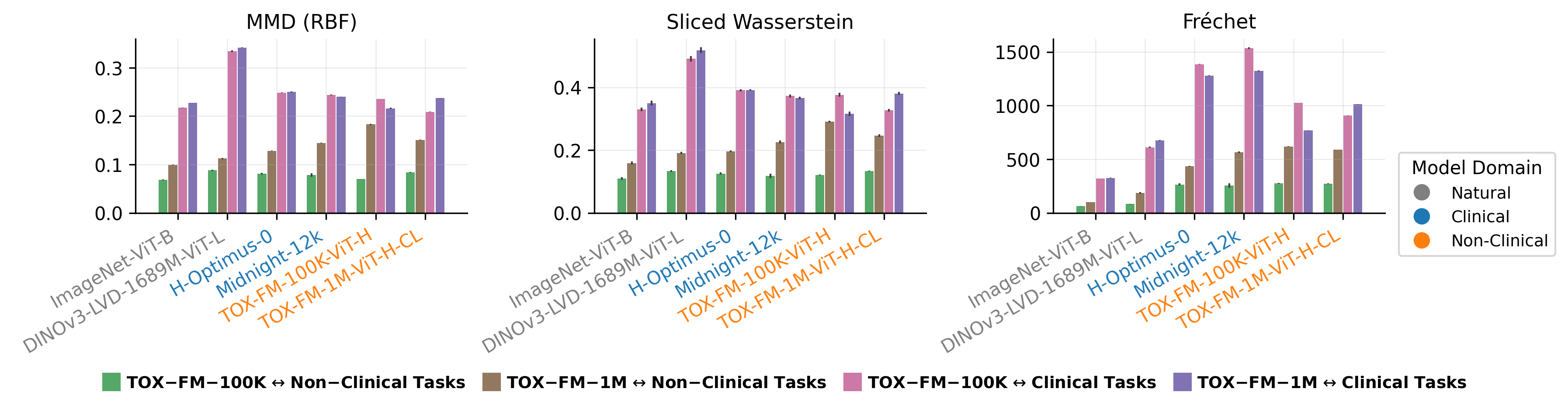}
\caption{Average pairwise distribution distances (mean and standard deviation) between the training and task datasets across three metrics.
Metrics are averaged across 5 random subsets of 100,000 patches from each dataset.}
\label{fig:figure-4}
\end{figure}

\textbf{Figure~\ref{fig:figure-4}} shows the pairwise distribution distances between the training and task datasets.
To ensure that the results were not biased by the choice of embedding space, we repeated the analysis using embeddings from six models across the natural image, clinical, and non-clinical domains.
Overall, the smallest distances across all metrics and embedding spaces are between \textbf{TOX-FM-100K} and the non-clinical tasks, consistent with our hypothesis.
Additionally, the distances between the training datasets and clinical tasks are considerably larger than those of non-clinical tasks, supporting the view that the clinical tasks constitute a challenging out-of-distribution benchmark for non-clinical PFMs.
UMAP visualizations of the datasets across all embedding spaces are provided in \hyperref[appendix-4]{\textbf{Appendix 4}}.

Therefore, while previous work has emphasized the positive effect of distributional proximity and demonstrated the limited effect of dataset diversity on downstream tasks beyond the training distribution \cite{RN56}, our findings suggest that both distributional proximity and dataset diversity may play different roles depending on the magnitude of the distribution shift.
In the low-shift regime, distributional proximity remains the dominant factor for performance on downstream tasks.
However, in the high-shift regime, we observe that training with the more diverse non-clinical \textbf{TOX-FM-1M} improves the performance on clinical tasks, suggesting that dataset diversity can enhance cross-domain performance even when distributional proximity is absent.

\begin{figure}[!htb]
\centering
\includegraphics[width=\linewidth]{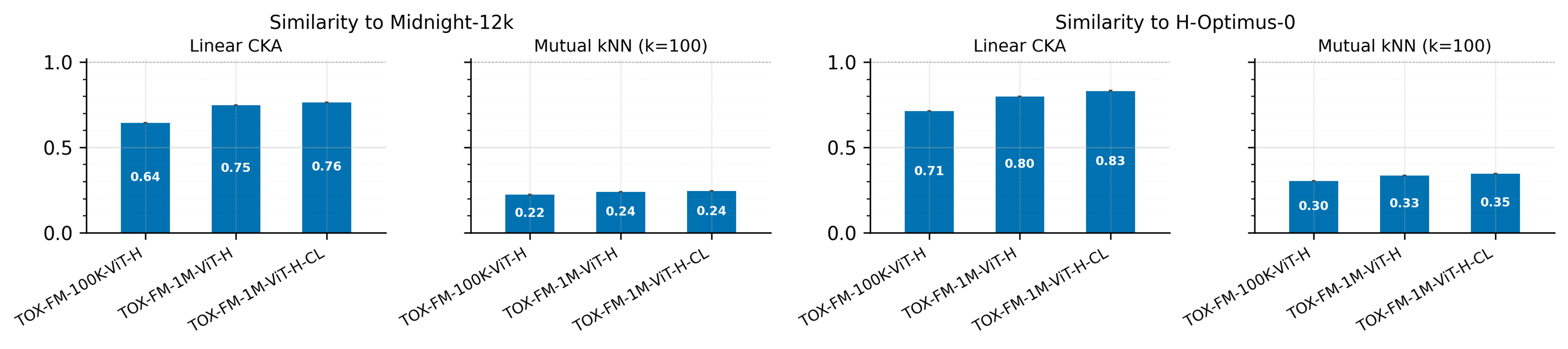}
\caption{Average representation similarity metrics (mean and standard deviation) of TOX-FM models with respect to Midnight-12k and H-optimus-0.
Metrics are averaged across 5 random subsets of 100,000 patches from the clinical task datasets.
We observed very small variability for both metrics across the random subsets.
We report the mutual kNN metric with k=100 but note that the same trend was observed for k=1 and k=10.}
\label{fig:figure-5}
\end{figure}

Motivated by recent work on the convergence of representations across models trained using different datasets, methods, and modalities \cite{RN61}, we further examined whether scaling dataset diversity is associated with the convergence of non-clinical and clinical PFMs, thereby providing a potential explanation for the observed cross-domain generalization of TOX-FM models.
Specifically, we measured the representation similarity of TOX-FM models with respect to Midnight-12k and H-optimus-0, the two best performing clinical PFMs on clinical tasks.
We calculated the centered kernel alignment (CKA) \cite{RN62} and mutual k-nearest neighbors (kNN) \cite{RN61} metrics with null calibration \cite{RN63}, which capture global and local representation similarity, respectively.
The metrics were calculated using random, stratified subsets of 100,000 patches from the clinical task datasets.
\textbf{Figure~\ref{fig:figure-5}} shows the representation similarity metrics between non-clinical and clinical PFMs.
Overall, both global and local representation similarities increase alongside training dataset diversity, highlighting the increased representation convergence between non-clinical and clinical PFMs and consistent with the overall best cross-domain performance demonstrated by TOX-FM-1M-ViT-H-CL.

\subsection{Continual Learning}\label{continual-learning}

\begin{figure}[!htb]
\centering

\begin{minipage}[c]{0.30\linewidth}
\vspace{0pt}
\centering

\includegraphics[width=\linewidth]{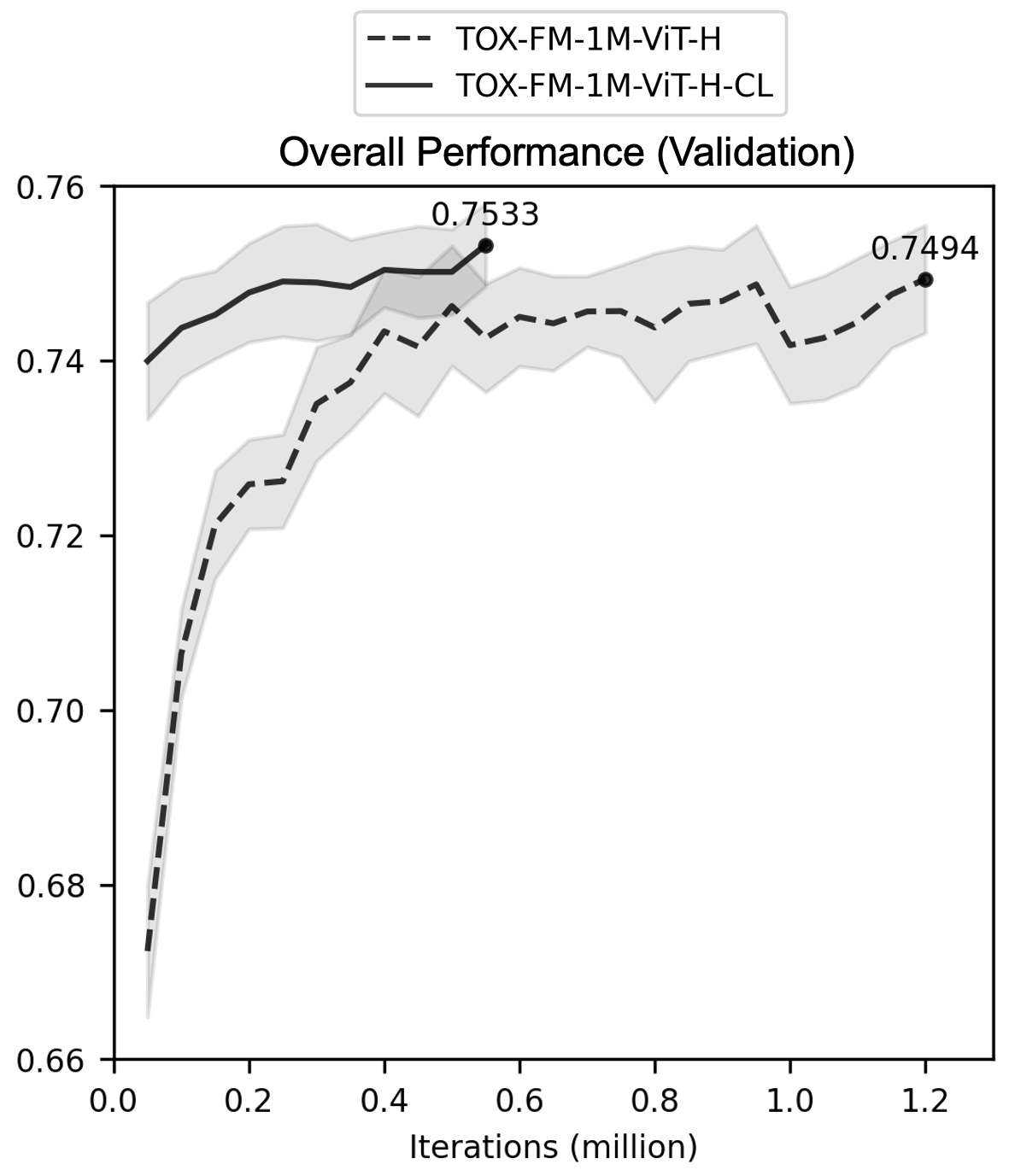}
\caption{Overall performance (mean and standard deviation) trajectories of TOX-FM-1M-ViT-H and TOX-FM-1M-ViT-H-CL on non-clinical validation downstream tasks during training on \textbf{TOX-FM-1M}.}
\label{fig:figure-6}
\end{minipage}
\hfill
\begin{minipage}[c]{0.65\linewidth}
\vspace{0pt}
\setlength{\parskip}{0.75em}
\setlength{\parindent}{0pt}

The continual pre-training strategy was found to be effective.
Specifically, TOX-FM-1M-ViT-H-CL outperformed both TOX-FM-100K-ViT-H (the initialization checkpoint) and TOX-FM-1M-ViT-H (the model trained from scratch) across downstream tasks, suggesting successful integration of knowledge from both datasets to yield the best overall performance among TOX-FM models.

\textbf{Figure~\ref{fig:figure-6}} shows the performance trajectories on the non-clinical validation downstream tasks from the initialization through the selected checkpoints during training on \textbf{TOX-FM-1M}.
By leveraging previous training efforts from the first phase on \textbf{TOX-FM-100K}, continual learning accelerated the convergence of TOX-FM-1M-ViT-H-CL and fewer iterations were required to achieve peak performance in the second phase, reflecting the reduced incremental computational costs of updating models as new datasets become available.

\end{minipage}
\end{figure}

\clearpage
\subsection{Gram Anchoring}\label{gram-anchoring}

\begin{figure}[!htb]
\centering

\begin{minipage}[c]{0.60\linewidth}
\vspace{0pt}
\setlength{\parskip}{0.75em}
\setlength{\parindent}{0pt}

\textbf{Figure~\ref{fig:figure-7}} shows the overall performance of TOX-FM models and the corresponding -Gram models on spatially resolved non-clinical and clinical downstream tasks.
More details of the performance on individual tasks are provided in \hyperref[appendix-5]{\textbf{Appendix 5}}.

Overall, Gram anchoring demonstrated modest performance gains on spatially resolved downstream tasks, and benefits were more consistent across clinical tasks than non-clinical tasks.
In particular, the comparison between TOX-FM-1M-ViT-H-CL and TOX-FM-1M-ViT-H-CL-Gram showed mixed results.
For example, while Gram anchoring significantly improved performance on the artifact patch classification task using linear probe evaluation, the performance decreased using attention probe evaluation.
Similarly, Gram anchoring improved performance on clinical segmentation tasks, but the same effects were not consistently observed for non-clinical segmentation tasks.

\textbf{Figure~\ref{fig:figure-8}} shows examples of the PCA feature maps of TOX-FM-1M-ViT-H-CL and TOX-FM-1M-ViT-H-CL-Gram.
More examples are provided in \hyperref[appendix-5]{\textbf{Appendix 5}}.
The PCA feature maps are visualized at different magnifications and fields of view to compare the effects of Gram anchoring across different scales.
Qualitatively, Gram anchoring feature maps seem to exhibit enhanced semantic coherence and reduced noise, especially at lower magnifications where spatial patterns of tissue structures are easier to discern visually.
While the differences between the PCA feature maps are more difficult to perceive at higher magnifications, the impact on the performance of downstream tasks suggest that the effects of Gram anchoring persist across all scales.

However, further investigations are needed to better understand the effects of Gram anchoring on the dense features and better identify the types of downstream tasks that may benefit from this approach, specifically for histopathology.
Additionally, further exploration of techniques to improve the training procedure, such as checkpoint selection of the Gram teacher, may be informative as well.

\end{minipage}
\hfill
\begin{minipage}[c]{0.35\linewidth}
\vspace{0pt}
\centering

\includegraphics[width=\linewidth]{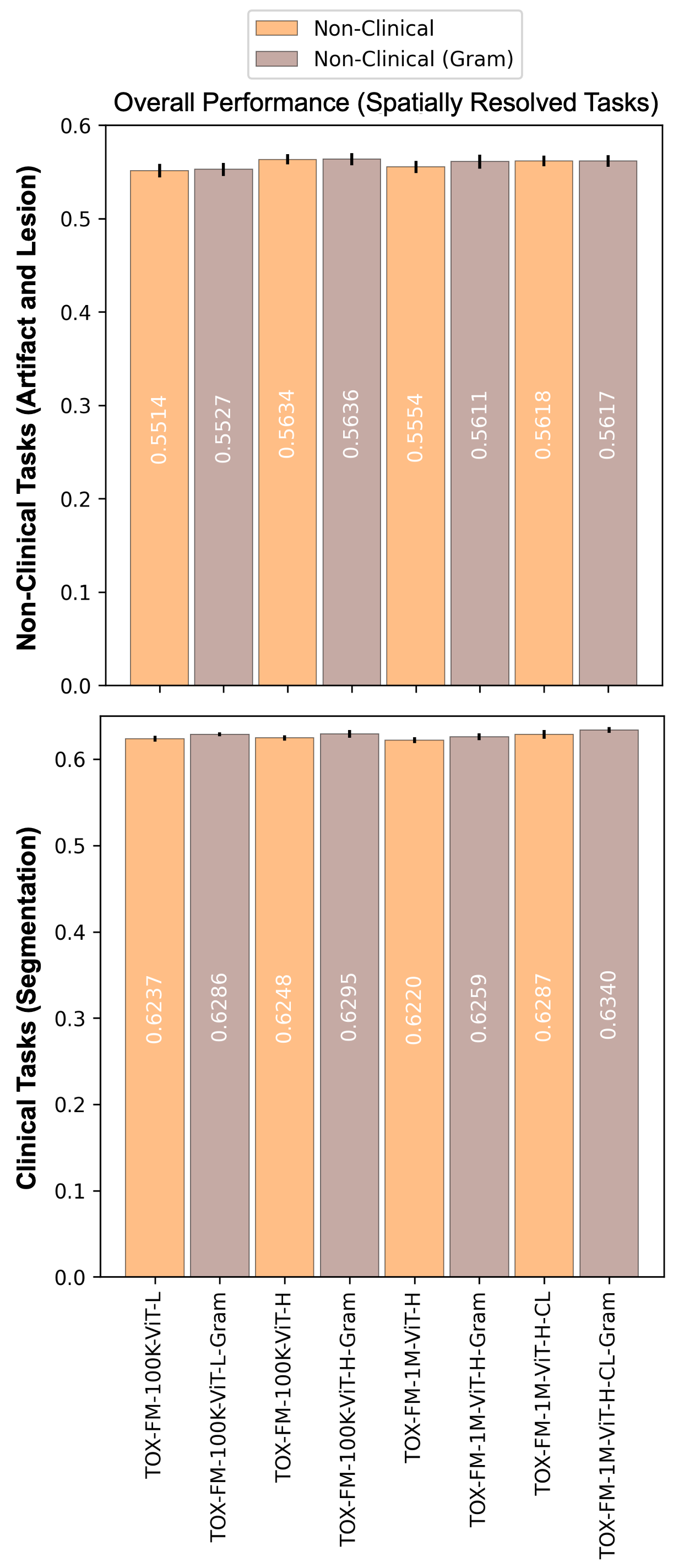}
\caption{Overall performance (mean and standard deviation) of TOX-FM models and the corresponding -Gram models on spatially resolved non-clinical (artifact and lesion) and clinical (segmentation) downstream tasks.}
\label{fig:figure-7}

\end{minipage}
\end{figure}

\begin{figure}[!htb]
\centering
\includegraphics[width=\linewidth]{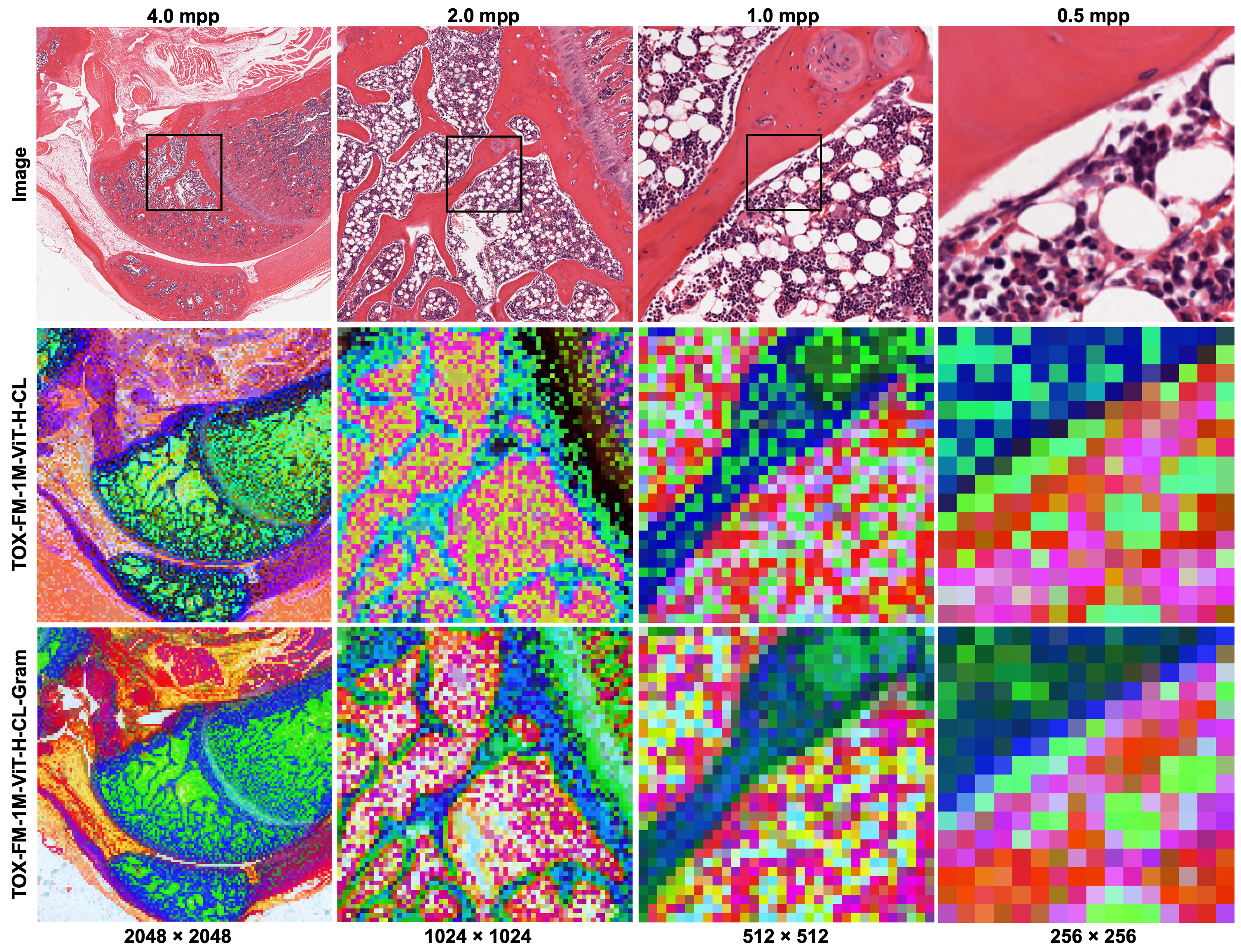}
\caption{Examples of PCA feature maps across magnifications and concentric fields of view on an H\&E WSI of femur tissue from a rat.
The black boxes on each image indicate the field of view at the next higher magnification.}
\label{fig:figure-8}
\end{figure}

\section{Discussion}\label{discussion}

We present TOX-FM, a family of non-clinical PFMs trained using the DINOv3 self-supervised learning framework on large-scale diverse histopathology datasets spanning a broad range of animal toxicology studies.
We evaluate TOX-FM and state-of-the-art clinical PFMs on a comprehensive set of 18 non-clinical and 9 clinical downstream tasks, addressing key open questions regarding the role of domain-specific training and the capability of cross-domain generalization of PFMs between human and animal tissues.

TOX-FM models showed the best performance on non-clinical downstream tasks, while clinical PFMs showed the best performance on clinical downstream tasks.
This advantage of domain-specific training is further emphasized by the computational efficiency of the ViT-L and ViT-H TOX-FM models, which demonstrate superior performance on non-clinical tasks with only 30 -- 60\% of the model capacity of ViT-g clinical PFMs.
While domain-specific models remain optimal, the cross-domain generalization capabilities of both non-clinical and clinical PFMs are notable, although transferability varies based on the type of tasks and the magnification at which the tasks are performed.
Further evaluation across both non-clinical and clinical domains, including broadening the scope and complexity of tasks across species, indications, tissues, and magnifications would be important to better understand where domain-specific models are necessary and where cross-domain models may be a viable alternative if domain-specific options are unavailable.
We also highlight the importance of dataset diversity, which exhibited a stronger impact than model size, especially on cross-domain performance.

To the best of our knowledge, we have provided the first analysis of cross-domain performance of PFMs across the non-clinical and clinical domains.
The evidence of bidirectional cross-domain generalization observed in our evaluations suggests that PFMs learn domain-invariant representations that capture fundamental cellular or morphological features that are conserved across species.
This could be particularly valuable in translational research, where a major challenge is the limited translatability of preclinical findings to clinical outcomes, resulting in many drug candidates failing to reach patients.
By providing a potentially transferable representation space, non-clinical PFMs have the potential to bridge this gap and support the discovery and development of more robust predictors of clinical outcomes at earlier stages of the process.

We also highlight the impact of distributional proximity and dataset diversity on downstream performance.
In line with previous work \cite{RN56}, we show that distributional proximity between the training and task datasets appears to be the dominant factor for performance on downstream tasks in the low-shift regime.
However, in contrast to previous work from the natural image domain \cite{RN56}, we show that dataset diversity can improve cross-domain performance in the high-shift regime in the histopathology domain, as demonstrated by the strong performance of non-clinical PFMs trained on more diverse non-clinical datasets on clinical downstream tasks.

TOX-FM-1M-ViT-H-CL emerged as the overall best TOX-FM model, leveraging a continual pre-training strategy to effectively integrate knowledge from datasets collected at two different phases while maintaining efficient use of computational resources.
We plan to use this scalable framework to continue incorporating new datasets as additional WSIs become available from future toxicology studies.
More broadly, this highlights the value of continual learning approaches, which have become increasingly important amid the rapid growth of data and the expanding scale of foundation model development.
Recent studies have proposed and evaluated a range of continual learning paradigms, focusing on the critical balance between plasticity and stability, as well as demonstrating the adaptability, efficiency, and sustainability of these methods \cite{RN43,RN64}.

We investigated Gram anchoring to improve dense feature representations for spatially resolved downstream tasks, including classification tasks focusing on artifacts and lesions, which may be very small, as well as segmentation tasks.
Although we visually observed refinement of the dense feature representations with Gram anchoring, the impact on performance across the various non-clinical and clinical downstream tasks remains nuanced and task dependent.
Additional analysis would be informative and help provide clarity on where Gram anchoring offers the greatest benefits in histopathology applications, specifically.

Building on these advances, a promising next step is the development of PFMs trained on integrated non-clinical and clinical datasets.
By leveraging complementary information across domains, the current strengths of domain-specific models may be combined and support broader generalization.
We plan to further investigate this direction in future work.
One potential approach would be to curate harmonized datasets that include both non-clinical and clinical resources.
For example, clinical datasets such as The Cancer Genome Atlas (TCGA) \cite{RN65} could be combined with TOX-FM datasets, with the dataset curation and sampling strategy applied to ensure balanced contributions across both domains.
Alternatively, the continual learning approach could be extended across domains.
For example, TOX-FM models could be continually pre-trained on TCGA, or clinical PFMs could be further adapted using TOX-FM datasets.

While vision models continue to dominate the landscape of PFMs, multimodal foundation models developed by aligning histopathology with complementary modalities are emerging as a promising direction.
Vision-language PFMs integrate histopathology with associated text from scientific textbooks, research articles, clinical reports, online educational platforms, and disease ontologies \cite{RN66,RN67,RN68,RN69,RN70}.
Vision-omics PFMs combine histopathology with matched gene expression profiles from sequencing-based spatial transcriptomics technologies, such as Visium and Xenium \cite{RN71,RN72}.
However, most development and validation efforts remain focused on clinical applications, especially oncology, and further investigations of multimodal PFMs focused on the non-clinical domain remain an important area of research.
Notably, pathologists' reports in toxicology studies contain descriptions of animal pathology findings, the Standard for Exchange of Non-clinical Data provides a framework for controlled terminology, and existing datasets pairing histopathology with spatial transcriptomics include samples from animal species, primarily mouse \cite{RN73,RN74}.
Together, these resources may pave the way for more systematic development and validation of multimodal non-clinical PFMs, enabling the characterization of complex toxicity mechanisms across multiple biological scales and contexts.

In this work, we address a critical gap in the development of foundation models for histopathology through the introduction of TOX-FM, a family of non-clinical PFMs trained on large-scale diverse histopathology datasets of over one million WSIs consisting of over 60 organs across six animal species from over 14,000 toxicology studies across six decades.
As PFMs continue to advance, rigorous understanding of the strengths and limitations of both non-clinical and clinical models is essential to fully leverage these tools to accelerate drug development and enable effective translation from preclinical to clinical stages and beyond.
Towards this goal, TOX-FM establishes a benchmark for non-clinical PFMs while providing early insights into cross-domain generalization capabilities of existing clinical PFMs.

\section{Acknowledgements}\label{acknowledgements}

The authors thank all the toxicologic pathologists who provided annotation support for the downstream evaluation tasks.

\appendix
\setcounter{section}{0}
\renewcommand{\thesection}{Appendix \arabic{section}:}
\renewcommand{\thefigure}{A\arabic{figure}}
\renewcommand{\theHfigure}{A\arabic{figure}}
\setcounter{figure}{0}
\renewcommand{\thetable}{A\arabic{table}}
\renewcommand{\theHtable}{A\arabic{table}}
\setcounter{table}{0}

\clearpage
\section{Hierarchical K-means}\label{appendix-1}

The dataset curation method for \textbf{TOX-FM-1M} used consistent settings for hierarchical k-means across all magnifications.
To decrease storage and computational requirements, we used PCA to project the CLS token embeddings from Midnight-12k from 1536 into 256 dimensions for the highest magnification (0.5 mpp) and 512 for all other magnifications.
Then, we applied k-means clustering with 4 bottom-up levels of 100K, 10K, 1K, and 30 clusters, respectively.
At each magnification, the target subset size was determined based on the desired sampling ratio, which we set as 25\% of all candidate patches at the lowest magnification (4.0 mpp) and 50\% for all other magnifications.
Note that candidate patches at the lowest magnification consist of all non-overlapping patches from tissue regions from all WSIs, while candidate patches at each subsequent magnification are conditioned on selected patches at the preceding magnification.
The target subset size was recursively allocated across clusters in the hierarchy, beginning at the top-level of 30 clusters.

\begin{figure}[!htb]
\centering
\includegraphics[width=0.8\linewidth]{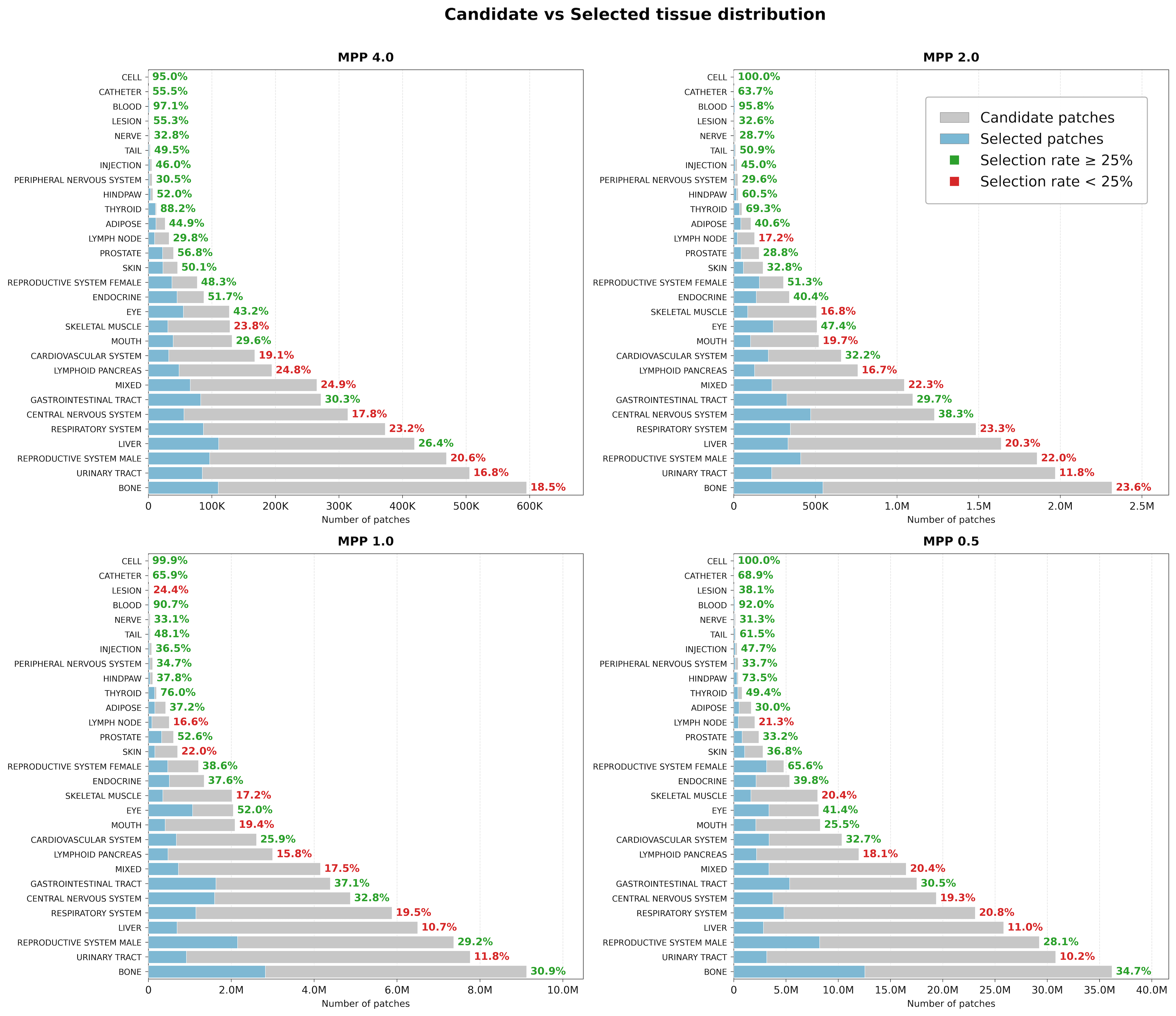}
\caption{Distribution of candidate and selected patches across tissue types using hierarchical k-means on a subset of 36,000 WSIs with corresponding tissue information.
For the validation analysis, the method was independently applied at each magnification with a fixed sampling ratio of 25\%.}
\label{fig:figure-a1}
\end{figure}

We performed a small-scale validation analysis on a subset of 36,000 rat WSIs for which corresponding tissue information was available.
For the validation analysis, we applied hierarchical k-means independently on all patches at each magnification (without applying the progressive multi-scale strategy) with a fixed sampling ratio of 25\%.
\textbf{Figure~\ref{fig:figure-a1}} shows the distribution of the candidate and selected patches across tissue types.
We observe that rare tissue types are generally oversampled (\(\geq\) 25\%) relative to the sampling ratio, while more common tissue types are generally sampled close to or slightly lower (\(<\) 25\%) than the sampling ratio.
Notably, while the method does not utilize the tissue metadata in any way, it produces clusters that seem to capture meaningful diversity across tissue phenotypes.

\clearpage
\section{Hyperparameters}\label{appendix-2}

\textbf{Tables~\ref{tab:table-a1}} and \textbf{\ref{tab:table-a2}} show the hyperparameter settings for non-clinical and clinical downstream tasks, respectively.

\small
\begin{longtable}{@{}
  >{\centering\arraybackslash}m{(\linewidth - 8\tabcolsep) * \real{0.20}}
  >{\centering\arraybackslash}m{(\linewidth - 8\tabcolsep) * \real{0.20}}
  >{\centering\arraybackslash}m{(\linewidth - 8\tabcolsep) * \real{0.20}}
  >{\centering\arraybackslash}m{(\linewidth - 8\tabcolsep) * \real{0.20}}
  >{\centering\arraybackslash}m{(\linewidth - 8\tabcolsep) * \real{0.20}}
@{}}
\toprule
\begin{minipage}[b]{\linewidth}\centering
\textbf{Component}
\end{minipage} & \begin{minipage}[b]{\linewidth}\centering
\textbf{Patch Classification (Linear Probe)}
\end{minipage} & \begin{minipage}[b]{\linewidth}\centering
\textbf{Patch Classification (Attention Probe)}
\end{minipage} & \begin{minipage}[b]{\linewidth}\centering
\textbf{Slide Classification}
\end{minipage} & \begin{minipage}[b]{\linewidth}\centering
\textbf{Segmentation}
\end{minipage} \\
\midrule\noalign{}
\endhead
\endlastfoot
Method & Logistic Regression & ABMIL & ABMIL & U-Net \\
Loss & Cross-entropy & Cross-entropy & Cross-entropy & Lovász \\
Learning rate & - & 0.01, 0.001, 0.0001 & 0.01, 0.003, 0.001 & 0.01,
0.001, 0.0001 \\
Scheduler & - & Cosine & Cosine & Cosine \\
Weight decay & - & 0.001, 0.0001 & 0.001, 0.0001 & 0.001, 0.0001 \\
Optimizer & L-BFGS & AdamW & AdamW & AdamW \\
Maximum iterations & 1000 & 5000 & 2500 & 5000 \\
Number of runs & 5 & 5 & 3 & 3 \\
\bottomrule
\caption{Hyperparameter settings for non-clinical downstream tasks.}
\label{tab:table-a1}
\end{longtable}
\normalsize

\small
\begin{longtable}{@{}
  >{\centering\arraybackslash}m{(\linewidth - 6\tabcolsep) * \real{0.20}}
  >{\centering\arraybackslash}m{(\linewidth - 6\tabcolsep) * \real{0.27}}
  >{\centering\arraybackslash}m{(\linewidth - 6\tabcolsep) * \real{0.27}}
  >{\centering\arraybackslash}m{(\linewidth - 6\tabcolsep) * \real{0.26}}
@{}}
\toprule
\begin{minipage}[b]{\linewidth}\centering
\textbf{Component}
\end{minipage} & \begin{minipage}[b]{\linewidth}\centering
\textbf{Patch Classification}
\end{minipage} & \begin{minipage}[b]{\linewidth}\centering
\textbf{Slide Classification}
\end{minipage} & \begin{minipage}[b]{\linewidth}\centering
\textbf{Segmentation}
\end{minipage} \\
\midrule\noalign{}
\endhead
\endlastfoot
Learning rate & 0.00003, \textbf{0.0003}, 0.003 & 0.0001,
\textbf{0.001}, 0.01 & 0.0002, \textbf{0.002}, 0.02 \\
Weight decay & 0.001, \textbf{0.01}, 0.1 & 0.001, \textbf{0.01}, 0.1 &
0.001, \textbf{0.01}, 0.1 \\
\bottomrule
\caption{Hyperparameter settings for the \emph{eva} evaluation framework for clinical downstream tasks.
The \textbf{bold} values indicate the defaults. All other hyperparameter settings followed the official \emph{eva} repository defaults.}
\label{tab:table-a2}
\end{longtable}
\normalsize

\clearpage
\section{Performance of Non-Clinical and Clinical PFMs}\label{appendix-3}

\textbf{Tables~\ref{tab:table-a3}} and\textbf{~\ref{tab:table-a4}} show the performance of all models on non-clinical and clinical downstream tasks, respectively.

\scriptsize
\begin{longtable}{@{}
  >{\centering\arraybackslash}m{(\linewidth - 20\tabcolsep) * \real{0.07}}
  >{\centering\arraybackslash}m{(\linewidth - 20\tabcolsep) * \real{0.10}}
  >{\centering\arraybackslash}m{(\linewidth - 20\tabcolsep) * \real{0.083}}
  >{\centering\arraybackslash}m{(\linewidth - 20\tabcolsep) * \real{0.097}}
  >{\centering\arraybackslash}m{(\linewidth - 20\tabcolsep) * \real{0.085}}
  >{\centering\arraybackslash}m{(\linewidth - 20\tabcolsep) * \real{0.085}}
  >{\centering\arraybackslash}m{(\linewidth - 20\tabcolsep) * \real{0.085}}
  >{\centering\arraybackslash}m{(\linewidth - 20\tabcolsep) * \real{0.10}}
  >{\centering\arraybackslash}m{(\linewidth - 20\tabcolsep) * \real{0.10}}
  >{\centering\arraybackslash}m{(\linewidth - 20\tabcolsep) * \real{0.095}}
  >{\centering\arraybackslash}m{(\linewidth - 20\tabcolsep) * \real{0.10}}
@{}}
\toprule
\begin{minipage}[b]{\linewidth}\centering
\textbf{Type}
\end{minipage} & \begin{minipage}[b]{\linewidth}\centering
\textbf{Task}
\end{minipage} & \begin{minipage}[b]{\linewidth}\centering
\textbf{ImageNet-ViT-B}
\end{minipage} & \begin{minipage}[b]{\linewidth}\centering
\textbf{DINOv3-LVD-1689M-ViT-L}
\end{minipage} & \begin{minipage}[b]{\linewidth}\centering
\textbf{Prov-GigaPath}
\end{minipage} & \begin{minipage}[b]{\linewidth}\centering
\textbf{H-optimus-0}
\end{minipage} & \begin{minipage}[b]{\linewidth}\centering
\textbf{Midnight-12k}
\end{minipage} & \begin{minipage}[b]{\linewidth}\centering
\textbf{TOX-FM-100K-ViT-L}
\end{minipage} & \begin{minipage}[b]{\linewidth}\centering
\textbf{TOX-FM-100K-ViT-H}
\end{minipage} & \begin{minipage}[b]{\linewidth}\centering
\textbf{TOX-FM-1M-ViT-H}
\end{minipage} & \begin{minipage}[b]{\linewidth}\centering
\textbf{TOX-FM-1M-ViT-H-CL}
\end{minipage} \\
\midrule\noalign{}
\endhead
\endlastfoot

\multirow[c]{8}{*}{%
    \raisebox{-6em}[0pt][0pt]{%
        \rotatebox[origin=c]{90}{%
            \parbox[c]{2.6cm}{\centering Patch Classification\\(Linear Probe)}%
        }%
    }%
} &
Species &
0.5627 (0.0019) &
0.6322 (0.0062) &
0.7461 (0.0046) &
0.7579 (0.0045) &
0.7746 (0.0062) &
0.8246 (0.0047) &
0.8345 (0.0066) &
0.8904 (0.0019) &
\textbf{0.8985 (0.0037)} \\
\cmidrule(lr){2-11} &
Tissue (Dog) &
0.8319 (0.0018) &
0.8596 (0.0042) &
0.9182 (0.0035) &
0.9136 (0.0011) &
0.9271 (0.0024) &
0.9363 (0.0031) &
0.9399 (0.0042) &
0.9610 (0.0022) &
\textbf{0.9633 (0.0027)} \\
\cmidrule(lr){2-11} &
Tissue (Monkey) &
0.9231 (0.0041) &
0.9373 (0.0045) &
0.9646 (0.0038) &
0.9604 (0.0037) &
0.9736 (0.0023) &
0.9783 (0.0023) &
0.9797 (0.0021) &
0.9840 (0.0022) &
\textbf{0.9865 (0.0020)} \\
\cmidrule(lr){2-11} &
Tissue (Mouse) &
0.7595 (0.0053) &
0.8215 (0.0052) &
0.8819 (0.0044) &
0.8770 (0.0041) &
0.8817 (0.0047) &
0.9099 (0.0070) &
\textbf{0.9158 (0.0053)} &
0.9092 (0.0057) &
0.9152 (0.0052) \\
\cmidrule(lr){2-11} &
Tissue (Rabbit) &
0.9292 (0.0029) &
0.9418 (0.0020) &
0.9636 (0.0029) &
0.9633 (0.0019) &
0.9648 (0.0024) &
0.9709 (0.0017) &
0.9710 (0.0023) &
0.9741 (0.0006) &
\textbf{0.9751 (0.0016)} \\
\cmidrule(lr){2-11} &
Tissue (Rat) &
0.8101 (0.0043) &
0.8435 (0.0020) & 
0.9163 (0.0053) &
0.9171 (0.0034) &
0.9173 (0.0043) &
0.9596 (0.0027) &
\textbf{0.9629 (0.0027)} &
0.9545 (0.0032) &
0.9578 (0.0028) \\
\cmidrule(lr){2-11} &
Artifact &
0.4944 (0.0058) &
0.5643 (0.0040) &
0.5987 (0.0042) &
0.5783 (0.0034) &
\textbf{0.6267 (0.0036)} &
0.6009 (0.0085) &
0.6112 (0.0049) &
0.5993 (0.0074) &
0.5916 (0.0046) \\
\cmidrule(lr){2-11} &
Lesion &
0.4556 (0.0096) &
0.4844 (0.0059) &
0.5850 (0.0054) &
\textbf{0.5927 (0.0069)} & 
0.5696 (0.0070) &
0.5657 (0.0093) &
0.5889 (0.0072) &
0.5732 (0.0055) &
0.5839 (0.0057) \\
\midrule

\multirow[c]{2}{*}{%
    \raisebox{-2em}[0pt][0pt]{%
        \rotatebox[origin=c]{90}{%
            \parbox[c]{2.6cm}{\centering Patch Classification\\(Attention Probe)}%
        }%
    }%
} &
\rule{0pt}{4em}\raisebox{1.5em}[0pt][0pt]{Artifact} &
0.6213 (0.0084) &
0.6699 (0.0051) &
0.7342 (0.0032) &
0.7311 (0.0055) &
0.7395 (0.0037) &
0.7380 (0.0066) &
0.7455 (0.0039) &
0.7435 (0.0052) &
\textbf{0.7529 (0.0047)} \\
\cmidrule(lr){2-11} &
\rule{0pt}{4em}\raisebox{1.5em}[0pt][0pt]{Lesion} &
0.5237 (0.0061) &
0.5029 (0.0021) &
0.6255 (0.0087) &
0.6394 (0.0055) &
0.6282 (0.0078) &
0.6392 (0.0082) &
0.6467 (0.0052) &
0.6374 (0.0063) &
\textbf{0.6582 (0.0029)} \\
\midrule

\multirow[c]{6}{*}{%
    \raisebox{-4em}[0pt][0pt]{%
        \rotatebox[origin=c]{90}{%
            Slide Classification
        }%
    }%
} &
Species &
0.8403 (0.0048) &
0.8343 (0.0080) &
0.9019 (0.0107) &
0.8986 (0.0033) &
0.9131 (0.0035) &
0.9672 (0.0042) &
0.9640 (0.0086) &
\textbf{0.9887 (0.0024)} &
0.9881 (0.0016) \\
\cmidrule(lr){2-11} &
Tissue (Dog) &
0.9749 (0.0078) &
0.9596 (0.0053) &
0.9781 (0.0040) &
0.9795 (0.0048) &
0.9888 (0.0041) &
0.9960 (0.0028) &
0.9953 (0.0025) &
0.9954 (0.0025) &
\textbf{0.9960 (0.0016)} \\
\cmidrule(lr){2-11} &
Tissue (Monkey) &
0.9436 (0.0061) &
0.9187 (0.0136) &
0.9388 (0.0102) &
0.9708 (0.0189) &
0.9987 (0.0009) &
0.9827 (0.0034) &
0.9778 (0.0027) &
\textbf{0.9993 (0.0009)} &
0.9953 (0.0041) \\
\cmidrule(lr){2-11} &
Tissue (Mouse) &
0.8894 (0.0078) &
0.8803 (0.0046) &
0.9351 (0.0028) &
0.9307 (0.0057) &
0.9439 (0.0063) &
0.9439 (0.0079) &
0.9515 (0.0080) &
0.9557 (0.0020) &
\textbf{0.9642 (0.0019)} \\
\cmidrule(lr){2-11} &
Tissue (Rabbit) &
0.9387 (0.0178) &
0.9387 (0.0135) &
0.9626 (0.0134) &
0.9572 (0.0103) &
0.9310 (0.0091) &
0.9960 (0.0043) &
0.9786 (0.0058) &
\textbf{0.9993 (0.0010)} &
0.9940 (0.0033) \\
\cmidrule(lr){2-11} &
Tissue (Rat) &
0.9601 (0.0016) &
0.9556 (0.0074) &
0.9709 (0.0035) &
0.9639 (0.0070) &
0.9779 (0.0072) &
0.9906 (0.0026) &
\textbf{0.9954 (0.0025)} &
0.9947 (0.0034) &
0.9907 (0.0008) \\
\midrule

\multirow[c]{2}{*}{%
    \raisebox{-2em}[0pt][0pt]{%
        \rotatebox[origin=c]{90}{%
            Segmentation
        }%
    }%
} &
\rule{0pt}{4em}\raisebox{1.5em}[0pt][0pt]{Artifact} &
0.2215 (0.0044) &
0.2389 (0.0098) &
0.2674 (0.0082) &
0.2540 (0.0060) &
0.2843 (0.0084) &
0.2796 (0.0065) &
\textbf{0.3015 (0.0021)} &
0.2950 (0.0061) &
0.2888 (0.0069) \\
\cmidrule(lr){2-11} &
\rule{0pt}{4em}\raisebox{1.5em}[0pt][0pt]{Lesion} &
0.3989 (0.0013) &
0.4068 (0.0016) &
0.4662 (0.0038) &
0.4795 (0.0061) &
0.4784 (0.0028) &
0.4852 (0.0020) &
0.4868 (0.0078) &
0.4838 (0.0082) &
\textbf{0.4955 (0.0078)} \\
\midrule

\multicolumn{2}{@{}>{\centering\arraybackslash}p{(\linewidth - 20\tabcolsep) * \real{0.1655} + 2\tabcolsep}}{Overall (Weighted)} &
0.5911 (0.0063) &
0.6145 (0.0063) &
0.6811 (0.0065) &
0.6818 (0.0059) &
0.6915 (0.0056) &
0.7033 (0.0063) &
0.7113 (0.0058) &
0.7157 (0.0053) &
\textbf{0.7205 (0.0047)} \\
\bottomrule
\caption{Average performance (mean and standard deviation) of all models on non-clinical downstream tasks.
Performance is reported based on macro F1 for classification tasks and mIOU for segmentation tasks.
Performance is averaged across multiple runs.
All tasks were assigned a weight of 1.0, except tissue tasks, which were assigned a weight of 0.2, to enable a more balanced contribution across task types and magnifications.}
\label{tab:table-a3}
\end{longtable}
\normalsize

\scriptsize
\begin{longtable}{@{}
  >{\centering\arraybackslash}m{(\linewidth - 20\tabcolsep) * \real{0.07}}
  >{\centering\arraybackslash}m{(\linewidth - 20\tabcolsep) * \real{0.10}}
  >{\centering\arraybackslash}m{(\linewidth - 20\tabcolsep) * \real{0.083}}
  >{\centering\arraybackslash}m{(\linewidth - 20\tabcolsep) * \real{0.097}}
  >{\centering\arraybackslash}m{(\linewidth - 20\tabcolsep) * \real{0.085}}
  >{\centering\arraybackslash}m{(\linewidth - 20\tabcolsep) * \real{0.085}}
  >{\centering\arraybackslash}m{(\linewidth - 20\tabcolsep) * \real{0.085}}
  >{\centering\arraybackslash}m{(\linewidth - 20\tabcolsep) * \real{0.10}}
  >{\centering\arraybackslash}m{(\linewidth - 20\tabcolsep) * \real{0.10}}
  >{\centering\arraybackslash}m{(\linewidth - 20\tabcolsep) * \real{0.095}}
  >{\centering\arraybackslash}m{(\linewidth - 20\tabcolsep) * \real{0.10}}
@{}}
\toprule
\begin{minipage}[b]{\linewidth}\centering
\textbf{Type}
\end{minipage} & \begin{minipage}[b]{\linewidth}\centering
\textbf{Task}
\end{minipage} & \begin{minipage}[b]{\linewidth}\centering
\textbf{ImageNet-ViT-B}
\end{minipage} & \begin{minipage}[b]{\linewidth}\centering
\textbf{DINOv3-LVD-1689M-ViT-L}
\end{minipage} & \begin{minipage}[b]{\linewidth}\centering
\textbf{Prov-GigaPath}
\end{minipage} & \begin{minipage}[b]{\linewidth}\centering
\textbf{H-optimus-0}
\end{minipage} & \begin{minipage}[b]{\linewidth}\centering
\textbf{Midnight-12k}
\end{minipage} & \begin{minipage}[b]{\linewidth}\centering
\textbf{TOX-FM-100K-ViT-L}
\end{minipage} & \begin{minipage}[b]{\linewidth}\centering
\textbf{TOX-FM-100K-ViT-H}
\end{minipage} & \begin{minipage}[b]{\linewidth}\centering
\textbf{TOX-FM-1M-ViT-H}
\end{minipage} & \begin{minipage}[b]{\linewidth}\centering
\textbf{TOX-FM-1M-ViT-H-CL}
\end{minipage} \\
\midrule\noalign{}
\endhead
\endlastfoot

\multirow[c]{5}{*}{%
    \raisebox{-3em}[0pt][0pt]{%
        \rotatebox[origin=c]{90}{%
            Patch Classification
        }%
    }%
} &
BACH &
0.6332 (0.0204) &
0.7164 (0.0048) &
0.7678 (0.0144) &
0.7704 (0.0102) &
\textbf{0.9051 (0.0043)} &
0.7158 (0.0053) &
0.7275 (0.0094) &
0.7973 (0.0119) &
0.8366 (0.0060) \\
\cmidrule(lr){2-11} &
BreakHis &
0.7271 (0.0073) &
0.7798 (0.0030) &
0.8348 (0.0123) &
0.8091 (0.0138) &
\textbf{0.8557 (0.0015)} &
0.7425 (0.0034) &
0.6385 (0.0043) &
0.7616 (0.0131) &
0.8059 (0.0033) \\
\cmidrule(lr){2-11} &
CRC &
0.9371 (0.0047) &
0.9278 (0.0023) &
0.9531 (0.0012) &
0.9562 (0.0037) &
0.9665 (0.0003) &
0.9577 (0.0009) &
0.9610 (0.0013) &
0.9710 (0.0007) &
\textbf{0.9739 (0.0009)} \\
\cmidrule(lr){2-11} &
Gleason Arvaniti &
0.6894 (0.0096) &
0.7451 (0.0051) &
0.7274 (0.0161) &
0.7709 (0.0052) &
\textbf{0.7991 (0.0021)} &
0.7469 (0.0027) &
0.7456 (0.0050) &
0.6995 (0.0103) &
0.6992 (0.0114) \\
\cmidrule(lr){2-11} &
Patch Camelyon &
0.8534 (0.0019) &
0.8856 (0.0023) &
\textbf{0.9450 (0.0008)} &
0.9421 (0.0010) &
0.9305 (0.0016) &
0.8874 (0.0029) &
0.8887 (0.0014) &
0.9131 (0.0006) &
0.9121 (0.0027) \\
\midrule

\multirow[c]{2}{*}{%
    \raisebox{-2em}[0pt][0pt]{%
        \rotatebox[origin=c]{90}{%
            Slide Classification
        }%
    }%
} &
\rule{0pt}{4em}\raisebox{1.5em}[0pt][0pt]{%
    \parbox[c]{1.5cm}{\centering Camelyon16\\(Small)}%
} &
0.6937 (0.0200) &
0.6577 (0.0613) &
0.8041 (0.0207) &
0.8269 (0.0253) &
\textbf{0.8459 (0.0213)} &
0.7803 (0.0191) &
0.7255 (0.0377) &
0.7501 (0.0670) &
0.7522 (0.0455) \\
\cmidrule(lr){2-11} &
\rule{0pt}{4em}\raisebox{1.5em}[0pt][0pt]{%
    \parbox[c]{1.5cm}{\centering PANDA\\(Small)}%
} &
0.4789 (0.0183) &
0.5551 (0.0089) &
0.6642 (0.0109) &
\textbf{0.6723 (0.0165)} &
0.6636 (0.0078) &
0.5845 (0.0056) &
0.5792 (0.0118) &
0.6032 (0.0102) &
0.6295 (0.0101) \\
\midrule

\multirow[c]{2}{*}{%
    \raisebox{-2em}[0pt][0pt]{%
        \rotatebox[origin=c]{90}{%
            Segmentation
        }%
    }%
} &
\rule{0pt}{4em}\raisebox{1.5em}[0pt][0pt]{BCSS} &
0.5660 (0.0081) &
0.5529 (0.0038) &
\textbf{0.6201 (0.0051)} &
0.6152 (0.0076) &
0.6098 (0.0037) &
0.6022 (0.0017) &
0.6034 (0.0033) &
0.6078 (0.0020) &
0.6070 (0.0029) \\
\cmidrule(lr){2-11} &
\rule{0pt}{4em}\raisebox{1.5em}[0pt][0pt]{MoNuSAC} &
0.6120 (0.0046) &
0.6209 (0.0045) &
0.6859 (0.0060) &
\textbf{0.6893 (0.0027)} &
0.6661 (0.0025) &
0.6452 (0.0046) &
0.6461 (0.0028) &
0.6363 (0.0039) &
0.6504 (0.0063) \\
\midrule

\multicolumn{2}{@{}>{\centering\arraybackslash}p{(\linewidth - 20\tabcolsep) * \real{0.1655} + 2\tabcolsep}}{Overall} &
0.6879 (0.0125) &
0.7157 (0.0209) &
0.7780 (0.0117) &
0.7836 (0.0121) &
\textbf{0.8047 (0.0079)} &
0.7403 (0.0073) &
0.7239 (0.0138) &
0.7489 (0.0237) &
0.7630 (0.0163) \\
\bottomrule
\caption{Average performance (mean and standard deviation) of all models on clinical downstream tasks.
Performance is reported based on metrics as defined in the official \emph{eva} repository.
Performance is averaged across multiple runs.}
\label{tab:table-a4}
\end{longtable}
\normalsize

\begin{figure}[!htb]
\centering
\includegraphics[width=\linewidth]{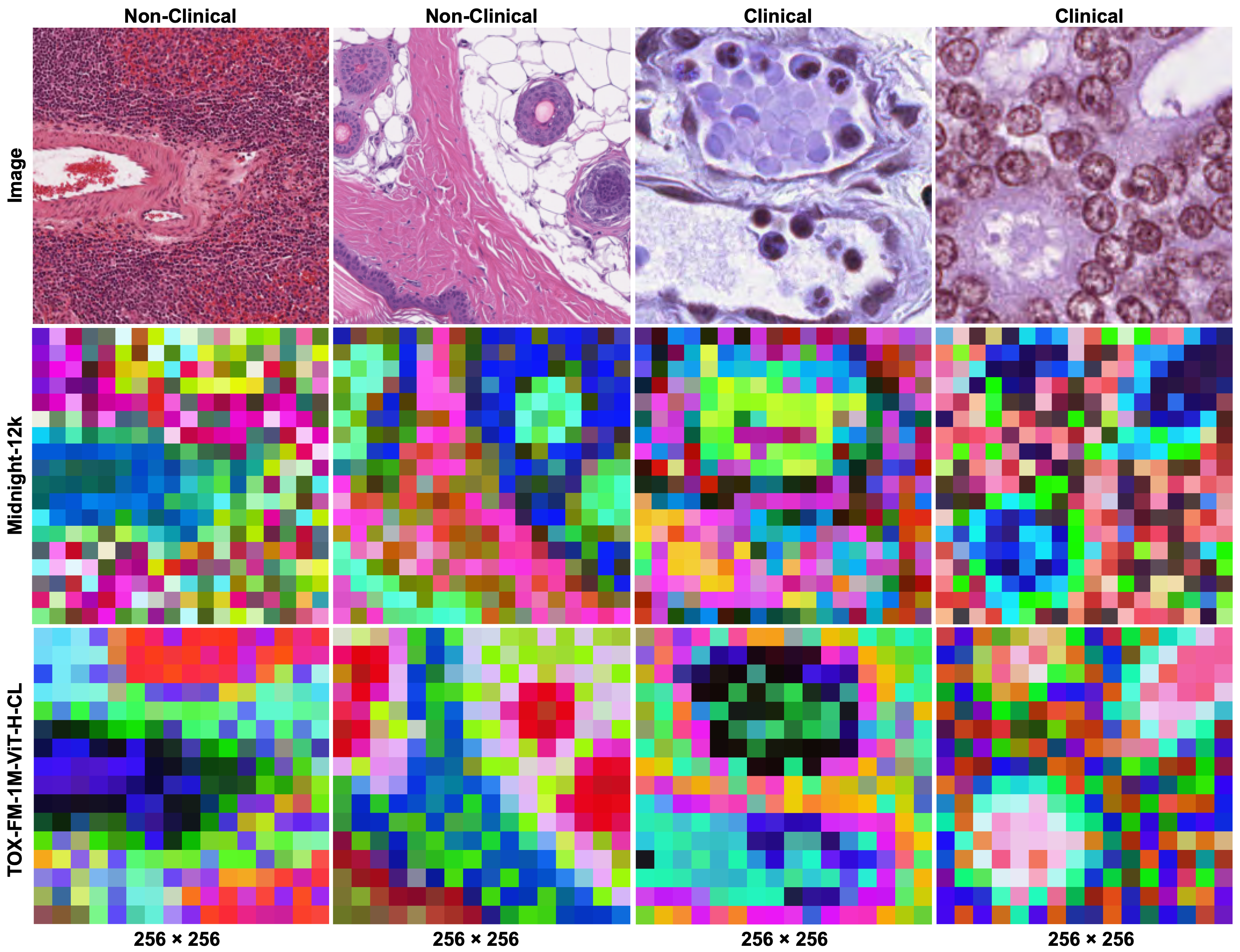}
\caption{Examples of PCA feature maps of Midnight-12k (best performing clinical PFM) and TOX-FM-1M-ViT-H-CL (best performing non-clinical PFM) across different domains at the training resolution.
Non-clinical images are taken from the tissue task (rat, 2.0 mpp) and clinical images are taken from the MoNuSAC task (0.25 mpp).}
\label{fig:figure-a2}
\end{figure}

\clearpage
\textbf{Figure~\ref{fig:figure-a2}} shows examples of PCA feature maps of Midnight-12k and TOX-FM-1M-ViT-H-CL at the training resolution of 256 × 256 pixels across the non-clinical and clinical domains.

\clearpage
\section{Dataset Diversity}\label{appendix-4}

\begin{figure}[!htb]
\centering
\includegraphics[width=\linewidth]{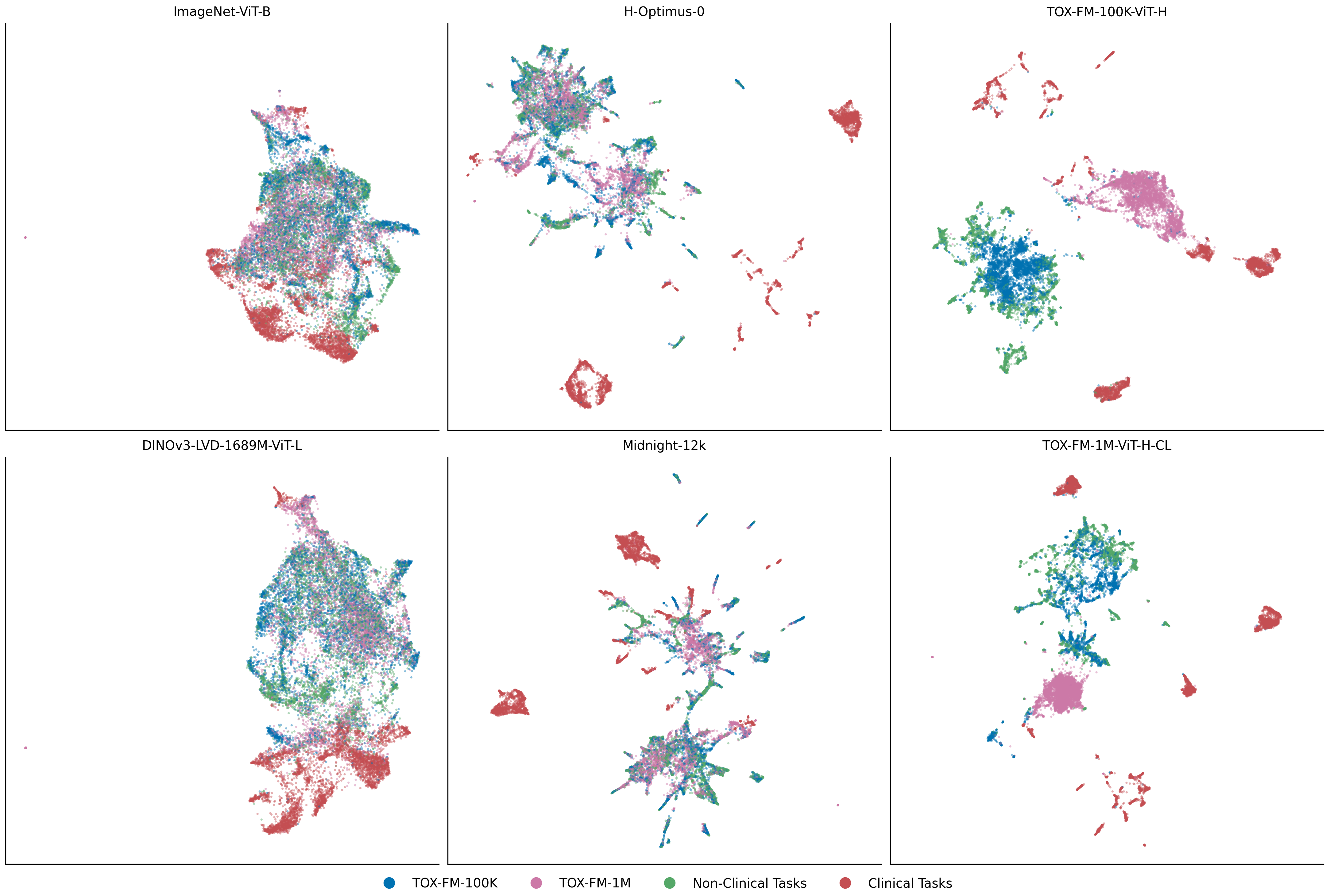}
\caption{UMAP visualizations of the training and task datasets across the embedding spaces of six models.
Each panel is fit independently on the respective model's embeddings.}
\label{fig:figure-a3}
\end{figure}

\textbf{Figure~\ref{fig:figure-a3}} shows the UMAP visualizations of the two training datasets (\textbf{TOX-FM-100K} and \textbf{TOX-FM-1M}) and the two task datasets (non-clinical and clinical) in the embedding spaces of six models across the natural image (ImageNet-ViT-B, DINOv3-LVD-1689M-ViT-L), clinical (H-optimus-0, Midnight-12k), and non-clinical (TOX-FM-100K-ViT-H, TOX-FM-1M-ViT-H-CL) domains.

All datasets collapse into an overlapping manifold for the natural image models.
In contrast, there is a clear separation of the clinical task datasets into several small clusters (red) across both clinical and non-clinical PFMs, reflecting the domain gap between clinical and non-clinical datasets.
Expectedly, the non-clinical task datasets (green) consistently fall within the training dataset distributions of \textbf{TOX-FM-100K} (blue) and \textbf{TOX-FM-1M} (pink), with an especially strong overlap observed with \textbf{TOX-FM-100K} in the embedding spaces of non-clinical PFMs, in line with our hypothesis regarding the distributional proximity of the datasets.

\clearpage
\section{Gram Anchoring}\label{appendix-5}

\textbf{Tables~\ref{tab:table-a5}} and\textbf{~\ref{tab:table-a6}} show the performance of TOX-FM models and the corresponding -Gram models on spatially resolved non-clinical (artifact and lesion) and clinical (segmentation) downstream tasks, respectively.

\scriptsize
\begin{longtable}{@{}
  >{\centering\arraybackslash}m{(\linewidth - 18\tabcolsep) * \real{0.040}}
  >{\centering\arraybackslash}m{(\linewidth - 18\tabcolsep) * \real{0.095}}
  >{\centering\arraybackslash}m{(\linewidth - 18\tabcolsep) * \real{0.105}}
  >{\centering\arraybackslash}m{(\linewidth - 18\tabcolsep) * \real{0.115}}
  >{\centering\arraybackslash}m{(\linewidth - 18\tabcolsep) * \real{0.105}}
  >{\centering\arraybackslash}m{(\linewidth - 18\tabcolsep) * \real{0.115}}
  >{\centering\arraybackslash}m{(\linewidth - 18\tabcolsep) * \real{0.105}}
  >{\centering\arraybackslash}m{(\linewidth - 18\tabcolsep) * \real{0.115}}
  >{\centering\arraybackslash}m{(\linewidth - 18\tabcolsep) * \real{0.105}}
  >{\centering\arraybackslash}m{(\linewidth - 18\tabcolsep) * \real{0.100}}
@{}}
\toprule
\begin{minipage}[b]{\linewidth}\centering
\textbf{Type}
\end{minipage} & \begin{minipage}[b]{\linewidth}\centering
\textbf{Task}
\end{minipage} & \begin{minipage}[b]{\linewidth}\centering
\textbf{TOX-FM-100K-ViT-L}
\end{minipage} & \begin{minipage}[b]{\linewidth}\centering
\textbf{TOX-FM-100K-ViT-L-Gram}
\end{minipage} & \begin{minipage}[b]{\linewidth}\centering
\textbf{TOX-FM-100K-ViT-H}
\end{minipage} & \begin{minipage}[b]{\linewidth}\centering
\textbf{TOX-FM-100K-ViT-H-Gram}
\end{minipage} & \begin{minipage}[b]{\linewidth}\centering
\textbf{TOX-FM-1M-ViT-H}
\end{minipage} & \begin{minipage}[b]{\linewidth}\centering
\textbf{TOX-FM-1M-ViT-H-Gram}
\end{minipage} & \begin{minipage}[b]{\linewidth}\centering
\textbf{TOX-FM-1M-ViT-H-CL}
\end{minipage} & \begin{minipage}[b]{\linewidth}\centering
\textbf{TOX-FM-1M-ViT-H-CL-Gram}
\end{minipage} \\
\midrule\noalign{}
\endhead
\endlastfoot

\multirow[c]{2}{*}{%
    \hspace{0.6em}%
    \raisebox{-2em}[0pt][0pt]{%
        \rotatebox[origin=c]{90}{%
            \parbox[c]{2.6cm}{\centering Patch Classification\\(Linear Probe)}%
        }%
    }%
} &
\rule{0pt}{4em}\raisebox{1.5em}[0pt][0pt]{Artifact} &
0.6009 (0.0085) &
0.6031 (0.0081) &
0.6112 (0.0049) &
\textbf{0.6163 (0.0084)} &
0.5993 (0.0074) &
0.6089 (0.0094) &
0.5916 (0.0046) &
0.6144 (0.0059) \\
\cmidrule(lr){2-10} &
\rule{0pt}{4em}\raisebox{1.5em}[0pt][0pt]{Lesion} &
0.5657 (0.0093) &
0.5681 (0.0080) &
0.5889 (0.0072) &
\textbf{0.5960 (0.0078)} &
0.5732 (0.0055) &
0.5828 (0.0086) &
0.5839 (0.0057) &
0.5807 (0.0080) \\
\midrule

\multirow[c]{2}{*}{%
    \hspace{0.6em}%
    \raisebox{-2em}[0pt][0pt]{%
        \rotatebox[origin=c]{90}{%
            \parbox[c]{2.6cm}{\centering Patch Classification\\(Attention Probe)}%
        }%
    }%
} &
\rule{0pt}{4em}\raisebox{1.5em}[0pt][0pt]{Artifact} &
0.7380 (0.0066) &
0.7391 (0.0066) &
0.7455 (0.0039) &
0.7483 (0.0051) &
0.7435 (0.0052) &
0.7505 (0.0056) &
\textbf{0.7529 (0.0047)} &
0.7481 (0.0070) \\
\cmidrule(lr){2-10} &
\rule{0pt}{4em}\raisebox{1.5em}[0pt][0pt]{Lesion} &
0.6392 (0.0082) &
0.6410 (0.0048) &
0.6467 (0.0052) &
0.6477 (0.0078) &
0.6374 (0.0063) &
0.6551 (0.0046) &
\textbf{0.6582 (0.0029)} &
0.6463 (0.0056) \\
\midrule

\multirow[c]{2}{*}{%
    \hspace{0.8em}%
    \raisebox{-2em}[0pt][0pt]{%
        \rotatebox[origin=c]{90}{%
            Segmentation%
        }%
    }%
} &
\rule{0pt}{4em}\raisebox{1.5em}[0pt][0pt]{Artifact} &
0.2796 (0.0065) &
0.2763 (0.0101) &
\textbf{0.3015 (0.0021)} &
0.2862 (0.0027) &
0.2950 (0.0061) &
0.2886 (0.0101) &
0.2888 (0.0069) &
0.2924 (0.0016) \\
\cmidrule(lr){2-10} &
\rule{0pt}{4em}\raisebox{1.5em}[0pt][0pt]{Lesion} &
0.4852 (0.0020) &
0.4884 (0.0027) &
0.4868 (0.0078) &
0.4873 (0.0037) &
0.4838 (0.0082) &
0.4809 (0.0044) &
\textbf{0.4955 (0.0078)} &
0.4886 (0.0081) \\
\midrule

\multicolumn{2}{@{}>{\centering\arraybackslash}p{(\linewidth - 18\tabcolsep) * \real{0.1305} + 2\tabcolsep}}{Overall} &
0.5514 (0.0073) &
0.5527 (0.0071) &
0.5634 (0.0055) &
\textbf{0.5636 (0.0063)} &
0.5554 (0.0065) &
0.5611 (0.0075) &
0.5618 (0.0057) &
0.5617 (0.0064) \\
\bottomrule
\caption{Average performance (mean and standard deviation) of TOX-FM models and the corresponding -Gram models on artifact and lesion non-clinical downstream tasks.
Performance is reported based on macro F1 for classification tasks and mIOU for segmentation tasks.
Performance is averaged across multiple runs.}
\label{tab:table-a5}
\end{longtable}
\normalsize

\scriptsize
\begin{longtable}{@{}
  >{\centering\arraybackslash}m{(\linewidth - 18\tabcolsep) * \real{0.040}}
  >{\centering\arraybackslash}m{(\linewidth - 18\tabcolsep) * \real{0.095}}
  >{\centering\arraybackslash}m{(\linewidth - 18\tabcolsep) * \real{0.105}}
  >{\centering\arraybackslash}m{(\linewidth - 18\tabcolsep) * \real{0.115}}
  >{\centering\arraybackslash}m{(\linewidth - 18\tabcolsep) * \real{0.105}}
  >{\centering\arraybackslash}m{(\linewidth - 18\tabcolsep) * \real{0.115}}
  >{\centering\arraybackslash}m{(\linewidth - 18\tabcolsep) * \real{0.105}}
  >{\centering\arraybackslash}m{(\linewidth - 18\tabcolsep) * \real{0.115}}
  >{\centering\arraybackslash}m{(\linewidth - 18\tabcolsep) * \real{0.105}}
  >{\centering\arraybackslash}m{(\linewidth - 18\tabcolsep) * \real{0.100}}
@{}}
\toprule
\begin{minipage}[b]{\linewidth}\centering
\textbf{Type}
\end{minipage} & \begin{minipage}[b]{\linewidth}\centering
\textbf{Task}
\end{minipage} & \begin{minipage}[b]{\linewidth}\centering
\textbf{TOX-FM-100K-ViT-L}
\end{minipage} & \begin{minipage}[b]{\linewidth}\centering
\textbf{TOX-FM-100K-ViT-L-Gram}
\end{minipage} & \begin{minipage}[b]{\linewidth}\centering
\textbf{TOX-FM-100K-ViT-H}
\end{minipage} & \begin{minipage}[b]{\linewidth}\centering
\textbf{TOX-FM-100K-ViT-H-Gram}
\end{minipage} & \begin{minipage}[b]{\linewidth}\centering
\textbf{TOX-FM-1M-ViT-H}
\end{minipage} & \begin{minipage}[b]{\linewidth}\centering
\textbf{TOX-FM-1M-ViT-H-Gram}
\end{minipage} & \begin{minipage}[b]{\linewidth}\centering
\textbf{TOX-FM-1M-ViT-H-CL}
\end{minipage} & \begin{minipage}[b]{\linewidth}\centering
\textbf{TOX-FM-1M-ViT-H-CL-Gram}
\end{minipage} \\
\midrule\noalign{}
\endhead
\endlastfoot

\multirow[c]{2}{*}{%
    \hspace{0.8em}%
    \raisebox{-2em}[0pt][0pt]{%
        \rotatebox[origin=c]{90}{%
            Segmentation%
        }%
    }%
} &
\rule{0pt}{4em}\raisebox{1.5em}[0pt][0pt]{BCSS} &
0.6022 (0.0017) &
0.6064 (0.0020) &
0.6034 (0.0033) &
0.6020 (0.0044) &
0.6078 (0.0020) &
\textbf{0.6118 (0.0032)} &
0.6070 (0.0029) &
0.6105 (0.0023) \\
\cmidrule(lr){2-10} &
\rule{0pt}{4em}\raisebox{1.5em}[0pt][0pt]{MoNuSAC} &
0.6452 (0.0046) &
0.6508 (0.0025) &
0.6461 (0.0028) &
0.6569 (0.0043) &
0.6363 (0.0039) &
0.6399 (0.0047) &
0.6504 (0.0063) &
\textbf{0.6576 (0.0044)} \\
\midrule

\multicolumn{2}{@{}>{\centering\arraybackslash}p{(\linewidth - 18\tabcolsep) * \real{0.1305} + 2\tabcolsep}}{Overall} &
0.6237 (0.0035) &
0.6286 (0.0023) &
0.6248 (0.0031) &
0.6295 (0.0043) &
0.6220 (0.0031) &
0.6259 (0.0041) &
0.6287 (0.0049) &
\textbf{0.6340 (0.0035)} \\
\bottomrule
\caption{Average performance (mean and standard deviation) of TOX-FM models and the corresponding -Gram models on segmentation clinical downstream tasks.
Performance is reported based on metrics as defined in the official \emph{eva} repository.
Performance is averaged across multiple runs.}
\label{tab:table-a6}
\end{longtable}
\normalsize

\begin{figure}[!htb]
\centering
\includegraphics[width=\linewidth]{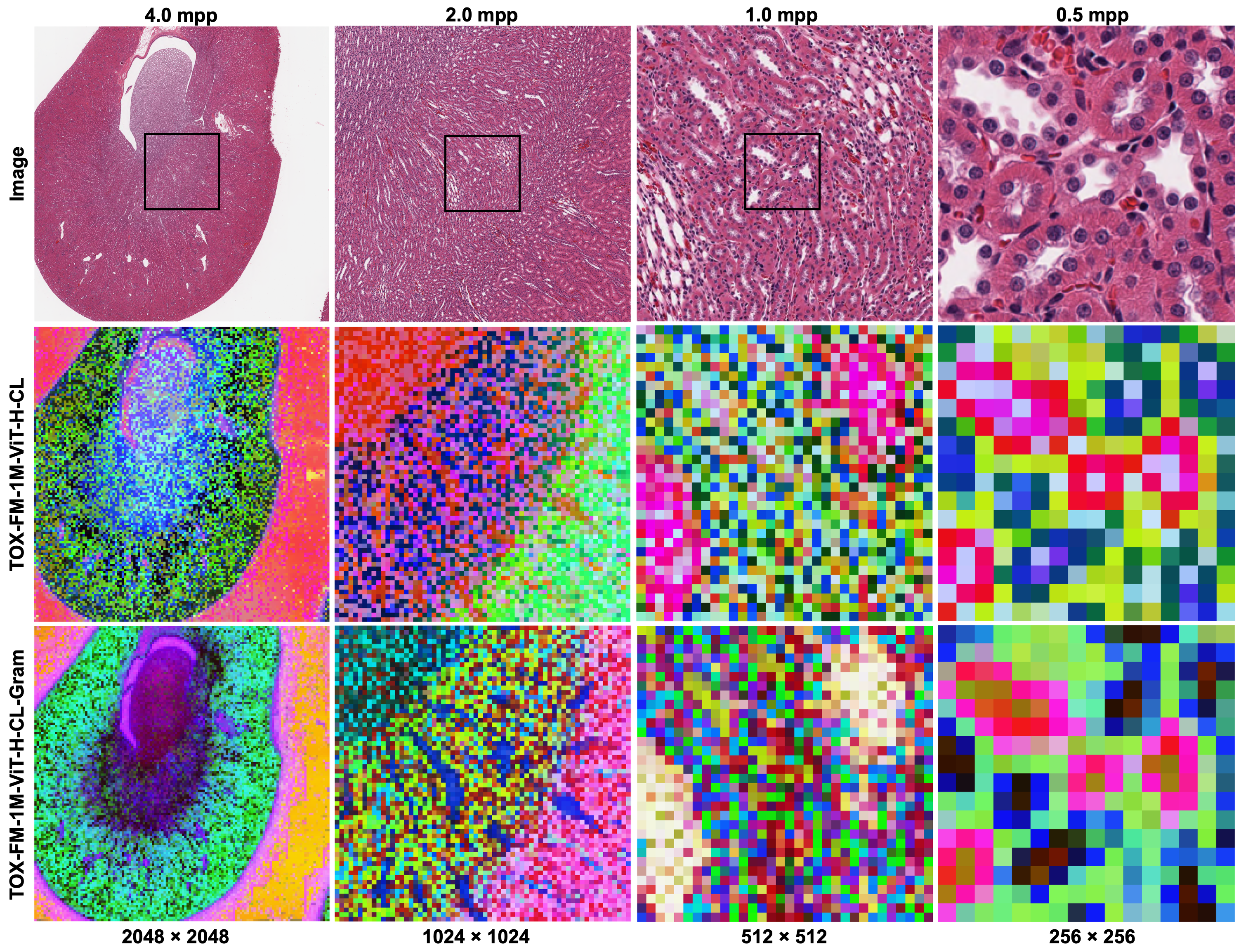}
\caption{Examples of PCA feature maps across magnifications and concentric fields of view on an H\&E WSI of kidney tissue from a rat.
The black boxes on each image indicate the field of view at the next higher magnification.}
\label{fig:figure-a4}
\end{figure}

\begin{figure}[!htb]
\centering
\includegraphics[width=\linewidth]{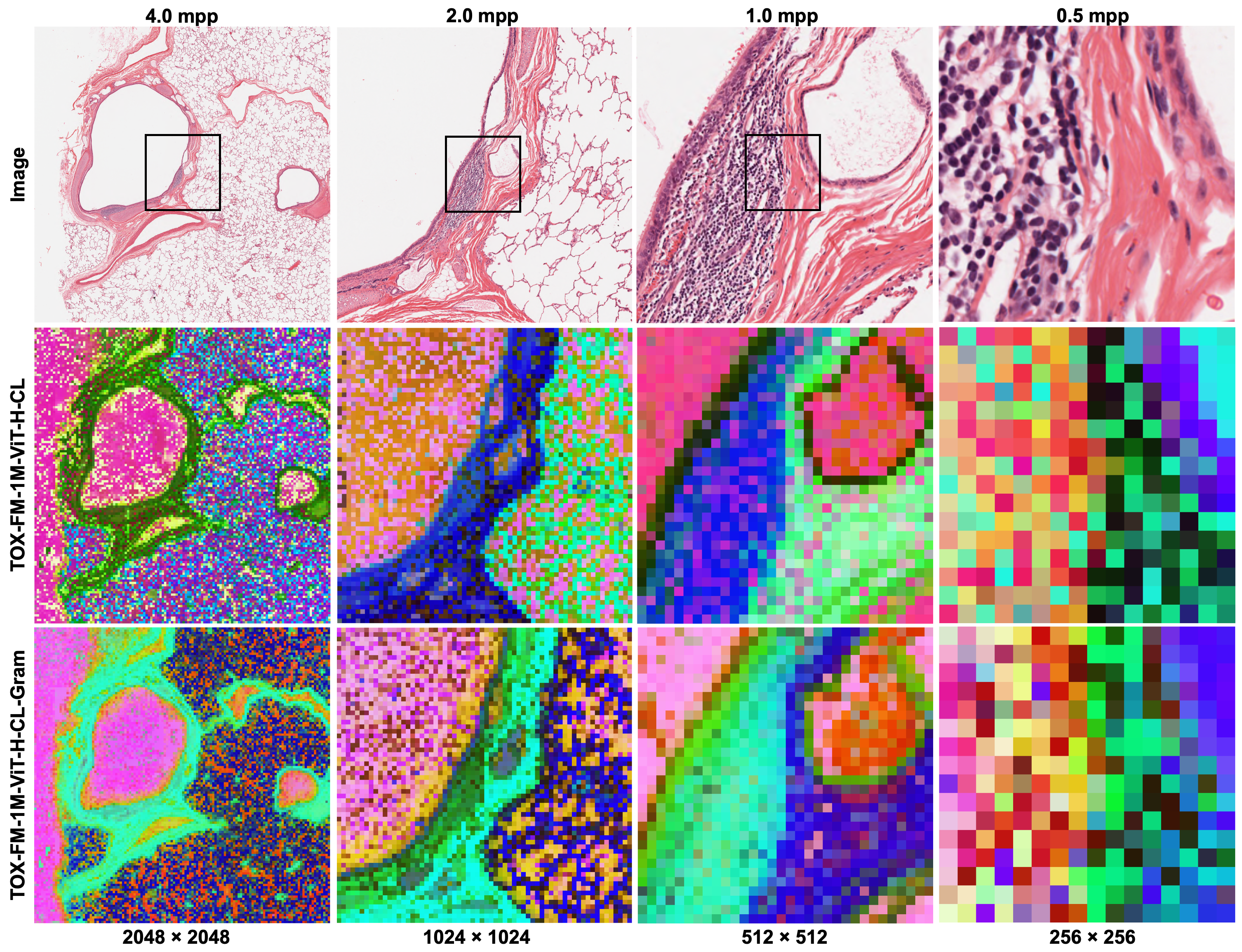}
\caption{Examples of PCA feature maps across magnifications and concentric fields of view on an H\&E WSI of lung tissue from a monkey.
The black boxes on each image indicate the field of view at the next higher magnification.}
\label{fig:figure-a5}
\end{figure}

\clearpage
\textbf{Figures~\ref{fig:figure-a4}} and \textbf{\ref{fig:figure-a5}} show examples of PCA feature maps across magnifications and fields of view for different tissues and species.

\clearpage
\bibliographystyle{unsrtnat}
\bibliography{references}

\end{document}